\documentclass[10pt,twocolumn,letterpaper]{article}
\usepackage[pagenumbers]{2025arXiv}
\definecolor{iccvblue}{rgb}{0.21,0.49,0.74}
\usepackage[pagebackref,breaklinks,colorlinks,allcolors=iccvblue]{hyperref}
\usepackage{amsmath, amsthm}
\theoremstyle{remark}
\newtheorem{theorem}{Theorem}[section]
\newtheorem{corollary}[theorem]{Corollary}
\usepackage{algorithm}
\usepackage{algpseudocode}
\theoremstyle{plain}
\usepackage{svg}
\usepackage{graphicx}

\makeatletter
\DeclareRobustCommand\onedot{\futurelet\@let@token\@onedot}
\def\@onedot{\ifx\@let@token.\else.\null\fi\xspace}
\def\eg{{e.g}\onedot}

\def\ie{{i.e}\onedot}

\def\etal{{et al}\onedot}
\makeatother
\newtheorem{assumptions}{Assumptions}
\newtheorem{proposition}{Proposition}

\usepackage{microtype}
\usepackage{linegoal}
\usepackage{multirow}
\usepackage{mathtools}
\usepackage{xspace}
\usepackage{bm}
\usepackage{amsthm}
\usepackage{booktabs}
\usepackage{nicefrac}
\newcommand{\CComments}[1]{{\color{DarkOrchid}#1}}
\newenvironment{sketchofproof}{\par\noindent{\bf Sketch of proof:\ }}{\hfill$\Box$\\\vspace{2mm}}

\title{Subspace-based Approximate Hessian Method for Zeroth-Order Optimization}

\author{
  Dongyoon~Kim \qquad\qquad Sungjae~Lee \qquad\qquad Wonjin~Lee \qquad\qquad Kwang~In~Kim\\
  POSTECH\\
  {\tt\small \{dykim910, leeeesj, wonjin0403, kimkin\}@postech.ac.kr}
}

\begin{document}

\maketitle
\begin{abstract}
Zeroth-order optimization addresses problems where gradient information is inaccessible or impractical to compute. While most existing methods rely on first-order approximations, incorporating second-order (curvature) information can, in principle, significantly accelerate convergence. However, the high cost of function evaluations required to estimate Hessian matrices often limits practical applicability. We present the subspace-based approximate Hessian (ZO-SAH) method, a zeroth-order optimization algorithm that mitigates these costs by focusing on randomly selected two-dimensional subspaces. Within each subspace, ZO-SAH estimates the Hessian by fitting a quadratic polynomial to the objective function and extracting its second-order coefficients. To further reduce function-query costs, ZO-SAH employs a periodic subspace-switching strategy that reuses function evaluations across optimization steps. Experiments on eight benchmark datasets, including logistic regression and deep neural network training tasks, demonstrate that ZO-SAH achieves significantly faster convergence than existing zeroth-order methods.
\end{abstract}

\section{Introduction}
\label{s:intro}
Many machine learning problems are formulated as optimization tasks, traditionally solved using gradient-based methods such as steepest descent~\cite{Bot12,Gar84,Ama93} and its variants~\cite{Zei15,DHS11,KB15}, which iteratively refine model parameters using first-order derivatives. However, in many real-world scenarios, gradients are either unavailable or impractical to compute. 

For example, in physics and chemistry applications, machine learning models often interact with non-differentiable simulators~\cite{LHC19,TKL22,TZF91,LCK20}. Similarly, in black-box learning settings~\cite{IEA18,ZYJ22,CZS17} including deep learning models integrated with third-party APIs, gradient access is restricted or entirely prohibited. Furthermore, specialized hardware implementations may not support conventional backpropagation~\cite{Ama93}. 

Zeroth-order (ZO) optimization offers a gradient-free alternative by approximating derivatives through finite differences~\cite{ZLH24,NS15,DJW15}. However, most existing ZO methods rely on first-order approximations. While incorporating second-order (curvature) information can, in principle, accelerate convergence~\cite{YHF18,ZDY25,Ye23}, its practical implementation is hindered by the high cost of estimating and inverting Hessian matrices. Traditional coordinate-wise finite difference methods require $O(d^2)$ function evaluations, where $d$ is the function's dimensionality, making them impractical for high-dimensional problems~\cite{NW06}.

To address this challenge, Ye~\etal proposed the ZO-Hessian aware (ZOHA) algorithm, which reduces function queries (evaluations) by estimating the Hessian using low-rank representations via approximate eigenvalue decomposition~\cite{Ye23}. While this \emph{power iteration}-based approach can theoretically reduce the function evaluation cost to $O(d)$ queries per iteration, achieving precise Hessian estimates in practice often still requires a substantial number of function evaluations~\cite{YHF18}.
Alternatively, Zhao~\etal introduced the diagonal Hessian informed ZO optimizer (HiZOO), designed for fine-tuning large language models. HiZOO improves query efficiency by approximating only the diagonal elements of the Hessian. While this strategy reduces function evaluation costs, it struggles with anisotropic functions due to its inability to capture parameter interdependence~(see~\cref{s:estimgradhess}).

In this paper, we present a new ZO algorithm called the subspace-based approximate Hessian (ZO-SAH) method. Our approach mitigates the high cost of Hessian estimation by focusing on randomly selected subspaces. At each parameter update step, ZO-SAH restricts optimization to a two-dimensional subspace, reducing both computational and function evaluation complexity to constant time.

Within each subspace, we estimate the Hessian by fitting a quadratic polynomial to the objective function and extracting its second-order coefficients. This method not only provides a more stable Hessian estimate than coordinate-wise finite differences but also enables the reuse of function evaluations from previous update steps, further enhancing efficiency.

Experiments on eight benchmark datasets show that ZO-SAH achieves significantly better convergence than existing ZO optimization methods in logistic regression and CNN-based image classification.

\section{Related work}
\paragraph{Zeroth-order optimization (ZOO).}
Most existing ZOO methods focus primarily on gradient estimation without leveraging second-order information. Broadly, these methods can be categorized into \emph{smoothing-based}~\cite{CLX19,GLZ16,LCC19} and (linear) \emph{interpolation-based}~\cite{BCC22,BCC19,KMR23,WQY24} approaches.

Smoothing-based methods estimate gradients by averaging multiple directional derivatives, typically sampling directions from the Gaussian distributions or uniform distribution on the unit sphere. For instance, ZO-signSGD uses only the sign of the estimated gradient components~\cite{LCC19}, while ZO-AdaMM improves convergence by integrating adaptive momentum techniques~\cite{CLX19}. These methods are effective when large sample sizes are feasible. However, their reliance on large samples limits practical applications where function queries are constrained due to computational or financial costs.

Interpolation-based methods estimate gradients by solving a linear system, often achieving more precise approximations than smoothing-based methods. Kozak~\etal employ finite differences along orthogonal random directions, a special case of linear interpolation~\cite{KMR23}. ReLIZO formulates gradient estimation as a quadratically constrained linear programming problem, deriving an analytical solution using Lagrange multipliers~\cite{WQY24}. 

While these methods improve gradient estimation, they do not incorporate curvature information, which could further enhance optimization efficiency.

\vspace{3mm}
\noindent\textbf{Hessian informed zeroth-order methods.\;}
Few studies have explored the incorporation of second-order information into ZO optimization. For example, ZOHA employs power iteration to estimate curvature~\cite{Ye23}. However, this approach requires $\mathcal{O}(d)$ function queries per iteration, with large constant factors, making ZOHA prohibitively expensive for high-dimensional optimization tasks.

To mitigate these costs, variations such as ZOHA-Gauss-DC and ZOHA-Diag-DC approximate the Hessian using a limited number of random directions~\cite{YHF18}. However, these methods lack formal complexity guarantees. HiZOO seeks to overcome this limitation using a \emph{mirror natural evolution} strategy but, due to high dimensionality, estimates only the diagonal components of the Hessian~\cite{ZDY25}. 

DeepZero~\cite{CZJ24} is among the first successful approaches for training deep neural networks using zeroth-order optimization, exploiting second-order information. It updates model parameters via coordinate-wise finite differences on randomly selected subspaces, guided by ZO-GraSP~\cite{CZJ24}, a zeroth-order adaptation of the GraSP neural network pruning algorithm~\cite{WZG20}. GraSP originally assigns pruning scores to optimization parameters based on the product of the Hessian and the gradient. 
In the zeroth-order setting, ZO-GraSP replaces this product with a finite-difference approximation computed from randomly perturbed parameter vectors. Since DeepZero is specifically designed for deep learning, its applicability to general zeroth-order optimization is limited. Our experiments show that a direct adaptation of DeepZero to broader ZO settings leads to significant performance degradation. In contrast, our ZO-SAH algorithm delivers competitive performance across a wider range of optimization problems, including deep learning.

\vspace{3mm}
\noindent\textbf{Sketching Newton-type method.\;}
The sketching Newton-type method is a variation of Newton’s method that reduces computational complexity by approximating the Hessian matrix using \emph{sketching techniques}, which apply randomized projections onto lower-dimensional subspaces to approximate the Hessian. This approach effectively reduces the problem’s dimensionality while preserving key curvature information, ultimately enabling more efficient Hessian inversion~\cite{GKL19,QRT16,Nes12,LAB16}.

Originally developed for conventional optimization, these algorithms require explicit computation of first- and second-order derivatives, making their direct extension to zeroth-order optimization nontrivial. Our work can be seen as a nontrivial adaptation of sketching Newton-type methods to the zeroth-order setting, with a particular emphasis on minimizing function query costs.

\section{Preliminaries}
\label{s:preliminarie}
\paragraph{First- and second-order optimization approaches.} Consider the problem of minimizing a twice-continuously differentiable function $f:\mathbb{R}^d\mapsto\mathbb{R}$. Two well-established strategies for this task are first-order and second-order iterative descent methods. At each iteration $k$, these methods update the current \emph{search point} $\mathbf{x}^k$ using the following update rules:
\begin{align}
\label{e:firstorderopt}
\mathbf{x}^{k+1}&=\mathbf{x}^k-\eta\nabla f(\mathbf{x}^k),&& \text{\small{(first-order)}}\\
\mathbf{x}^{k+1}&=\mathbf{x}^k-\eta(\mathbf{B}^k)^{-1}\nabla f(\mathbf{x}^k),&&\text{\small{(second-order)}}
\label{e:secondorderopt}
\end{align}
where $\eta$ is the step size (or learning rate), and $\mathbf{B}^k$ is a positive definite (PD) matrix guiding the search direction and scaling. A representative example of a second-order method is Newton's method, where $\mathbf{B}^k$ corresponds to the Hessian matrix $\mathbf{H}^k:=Hf(\mathbf{x}^k)$, with $H$ denoting the Hessian operator.

Second-order methods typically achieve faster convergence to local optima than first-order methods, both theoretically and empirically. However, for high-dimensional functions $f$, computing and inverting $\mathbf{B}^k$ can be prohibitively expensive. See~\cref{s:proposedmethod} for an illustration of Newton's method.

\vspace{3mm}
\noindent\textbf{Zeroth-order instantiations of first- and second-order optimization.\;}
When the objective function $f$ is not directly accessible and is instead presented as a black-box function, where $f(\mathbf{x})$ can be evaluated for any input $\mathbf{x}\in \mathbb{R}^d$ but direct access to its gradient $\nabla f$ and Hessian $\mathbf{B}$ is not available, zeroth-order (ZO) optimization offers a practical approach. In many such applications, evaluating $f$ incurs computational or financial costs, as seen in software systems that use commercial large language models offered as a service. In these scenarios, minimizing the number of function queries to reach the optimum becomes a critical performance criterion for an optimization algorithm. 

A common approach in ZO optimization is to approximate the gradient $\nabla f(\mathbf{x})$ using randomized finite difference methods while adhering to the first-order update rule (\cref{e:firstorderopt})~\cite{NS15,DJW15}:
\begin{equation}
\label{e:rge}
\widehat{\nabla}f(\mathbf{x}) = \frac{1}{q}\sum_{i=1}^{q} \frac{f(\mathbf{x}+\epsilon \mathbf{u}_{i})-f(\mathbf{x})}{\epsilon}\mathbf{u}_{i},
\end{equation}
where $\epsilon$ is a small positive scalar, $q$ is the number of perturbations, and $\mathbf{u}_i\in \mathbb{R}^d$ is a random perturbation vector drawn from the normal distribution $\mathcal{N}(\mathbf{0},\mathbf{I})$. It can be shown that $\widehat{\nabla}f(\mathbf{x})$ converges to the true gradient $\nabla f(\mathbf{x})$ as $q$ approaches infinity~\cite{NS15,DJW15}. In practice, the sample size $q$ significantly influences the quality of the gradient approximation $\widehat{\nabla}f$. Additional details on the derivations of $\widehat{\nabla}f$ can be found in~\cref{s:preliminaries}.

\section{Subspace-based approximate Hessians}
\label{s:proposedmethod}
Our approach, the subspace-based approximate Hessian (ZO-SAH), aims to enhance convergence in ZO optimization by translating the efficiency of second-order methods (\eg, Newton's method) into the ZO framework. A natural approach to achieving this is to replace the Hessian $\mathbf{B}$ in \cref{e:secondorderopt} with a finite-difference-based estimate~\cite{YHF18}:
\begin{align}
\label{e:hessiafd_supp}
\widehat{\mathbf{H}} = \frac{1}{q}\sum_{i=1}^{q}\frac{f(\mathbf{x}+\epsilon\mathbf{u}_i)+f(\mathbf{x}-\epsilon\mathbf{u}_i)-2f(\mathbf{x})}{2\epsilon^2}\mathbf{u}_{i}\mathbf{u}_{i}^{\top} + \lambda \mathbf{I},
\end{align}
where $\lambda$ is a regularization parameter that ensures $\widehat{\mathbf{H}}$ remains PD. This estimated Hessian can be combined with the finite-difference gradient $\widehat{\nabla}f$ (\cref{e:rge}) to enable second-order updates within the ZO setting.

However, for high-dimensional functions $f$, directly applying second-order updates (\cref{e:secondorderopt}) within a ZO framework introduces challenges beyond just computational and memory constraints. Accurate Hessian estimation in this setting requires a significantly large number of function queries, as the query complexity $q$ increases quadratically with the function dimensionality $d$, making direct application of second-order ZO optimization impractical in high-dimensional settings.

\subsection{Random subspace-based approach}
\label{s:randomsubspace}
To overcome these challenges, ZO-SAH adopts a random subspace-based strategy, which iteratively defines low-dimensional subspaces $\mathcal{W}$ and constructs an approximate Hessian therein. In the most extreme case, we restrict the dimensionality of $\mathcal{W}$ to two, reducing Hessian estimation to computing only three coefficients, given its symmetry.

While our approach is compatible with any method for generating $\mathcal{V}$, we adopt a simple two-step strategy in this work: First, we select an intermediate high-dimensional subspace $\mathcal{V}$ with an  (even) dimensionality $m$ by randomly choosing indices from $d$. Next, we \emph{cover} $\mathcal{V}$ by constructing a disjoint set of two-dimensional subspaces $\{\mathcal{W}_1,\ldots,\mathcal{W}_{\frac{m}{2}}\}$, each formed by pairing components from a random permutation of $\{1,\ldots,m\}$. This approach lies in-between purely random sampling techniques (\eg~\cref{e:hessiafd_supp}) and traditional coordinate descent approaches~\cite{NW06}.

At a given search point $\mathbf{x}$, suppose subspaces 
$\mathcal{V}$ and $\{\mathcal{W}_j\}_{j=1}^{\frac{m}{2}}$ are defined as described above. We parameterize $\mathbf{x}$ in $\mathcal{W}_j$ using a projection operator $P_j$, which is implemented through multiplication with an orthogonal matrix $\mathbf{P}_j$:
\begin{align}
\bm{\theta}_j = P_j(\mathbf{x}) = \mathbf{P}_j\mathbf{x}.
\label{e:projection}
\end{align}
The inverse mapping $P_j^{-1}(\bm{\theta}_j)$ is constructed by embedding the values of $\bm{\theta}_j$ back into the corresponding elements of $\mathbf{x}$, while keeping the remaining elements unchanged. 

Using this form, a single update step at iteration $k$ in our algorithm is achieved by accumulating these parameter updates across subspaces $\{\mathcal{W}_j\}$:
\begin{align}
\mathbf{x}^{k+1}&=\mathbf{x}^{k}-\eta \mathbf{v}^k,\quad \mathbf{v}^k=\sum_{j=1}^{\frac{m}{2}}P^{-1}\left((\widehat{\mathbf{H}}_j^k)^{-1}\widehat{\mathbf{g}}_j^k\right),
\label{e:ourupdate}
\end{align}
where $\widehat{\mathbf{g}}_j^k$ and $\widehat{\mathbf{H}}_j^k$ are the estimated gradient and Hessian at $\mathbf{x}^k$, respectively, restricted to the $\mathcal{W}_j$. For simplicity of notation, we omit the step index $k$ and interpret $f(\bm{\theta})$ as $f(\mathbf{x})$ in the subsequent discussion. 

\subsection{Estimating subspace gradient and Hessian}
\label{s:estimgradhess}
\paragraph{Gradient estimation.}
We estimate gradients using the standard coordinate-wise finite-difference~\cite{KMR23}: Within the given subspace $\mathcal{W}$, parameterized by $\bm{\theta}=[\theta^1,\theta^2]^\top\in\mathbb{R}^2$, the gradient estimate $\widehat{\mathbf{g}} =[\widehat{g}^1,\widehat{g}^2]^\top$ is computed as:
\begin{align}
\label{e:ourgradient}
\widehat{g}^1\approx \frac{f(\bm{\theta}_1)-f(\bm{\theta})}{\epsilon},
\end{align}
where $\bm{\theta}_1=\bm{\theta}+[\epsilon,0]^\top$, with a small $\epsilon$ (fixed at $10^{-3}$ in all experiments). The second component $\widehat{g}^2$ is computed analogously using $\bm{\theta}_2=\bm{\theta}+[0,\epsilon]^\top$. 

\begin{figure}
\centering
\footnotesize{
\begin{tabular}{c@{\hspace{1.0mm}}c@{\hspace{1.0mm}}l}
\includegraphics[width=0.495\columnwidth]{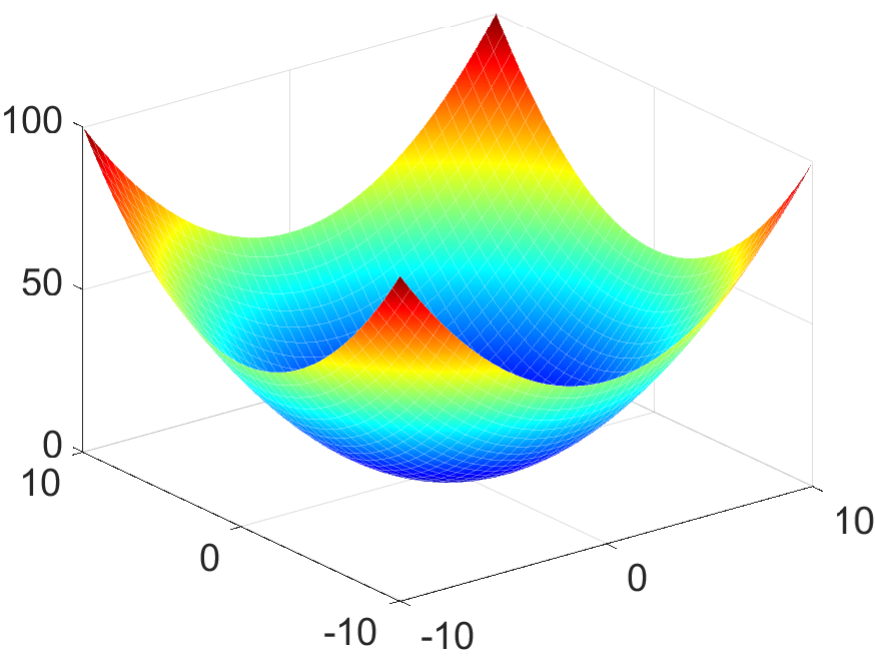}&
\includegraphics[width=0.495\columnwidth]{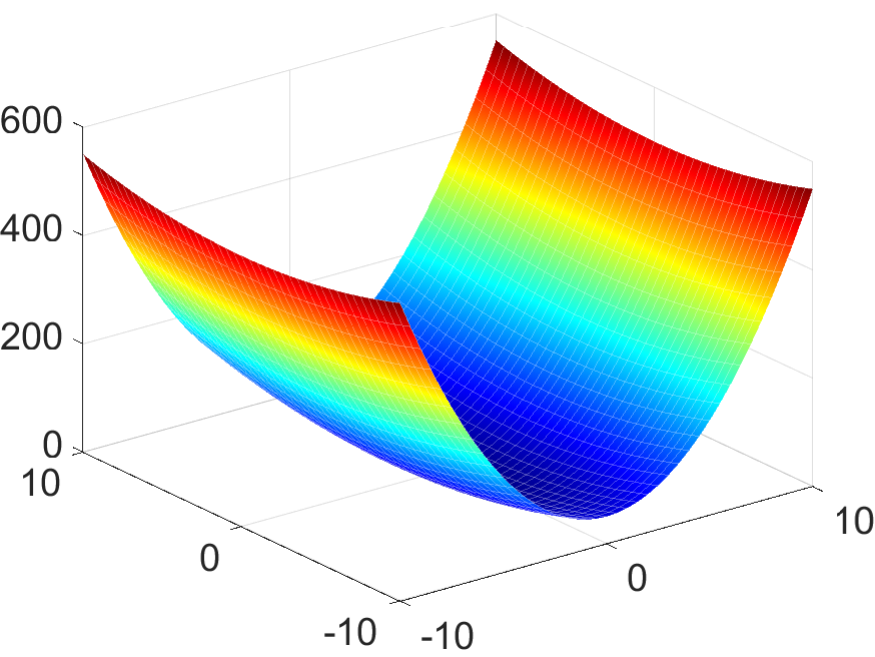}\\
(a) $\mathbf{A}=\begin{pmatrix} 1 & 0 \\ 0 & 1\end{pmatrix}$&
(b) $\mathbf{A}= \begin{pmatrix} 10 & 0 \\ 0 & 1\end{pmatrix}$\\
\includegraphics[width=0.495\columnwidth]{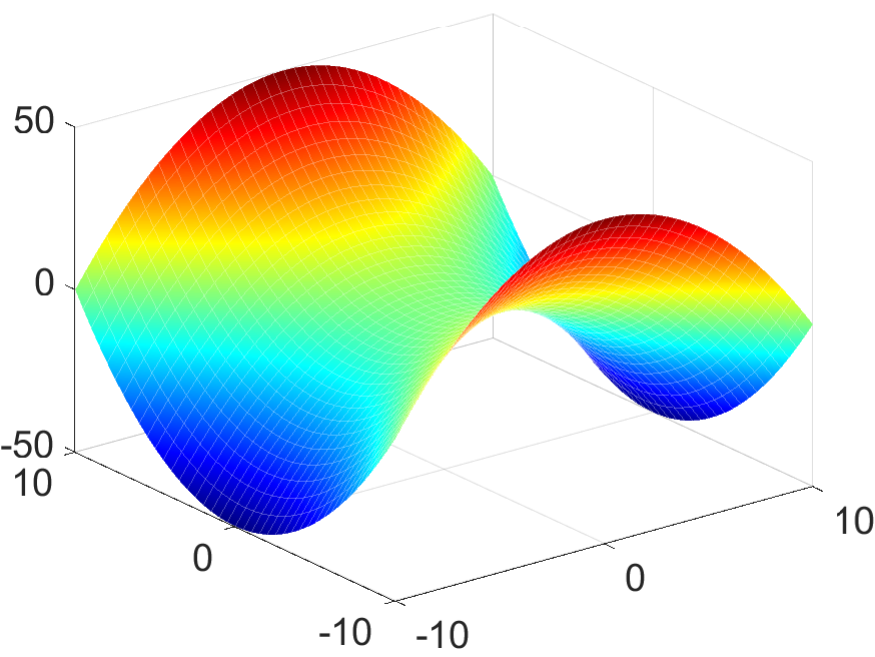}&
\includegraphics[width=0.495\columnwidth]
{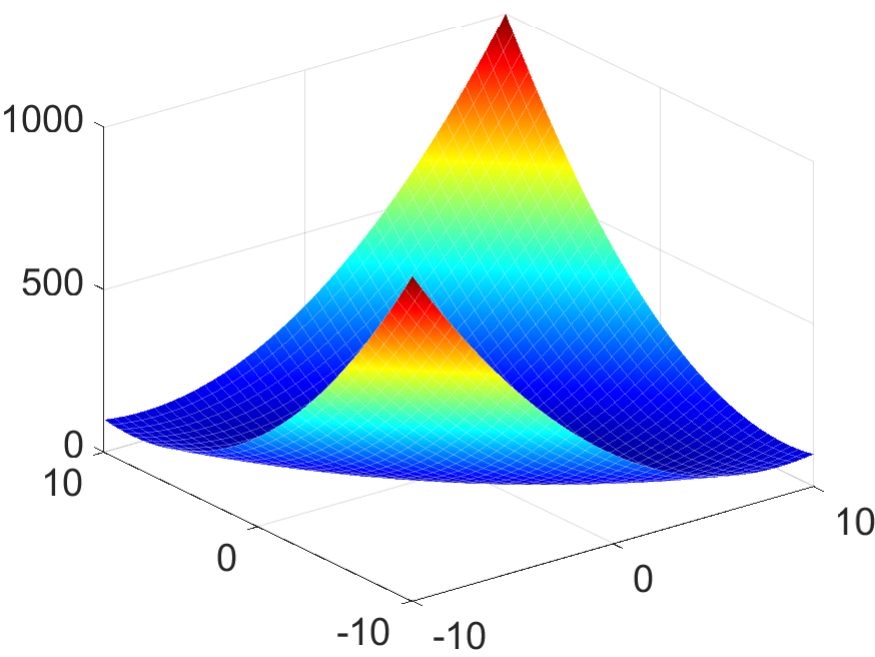}\\
(c) $\mathbf{A}=\begin{pmatrix} -1 & 0 \\ 0 & 1\end{pmatrix}$&
(d) $\mathbf{A}= \begin{pmatrix} 5.5 & 4.5 \\ 4.5 & 5.5\end{pmatrix}$\vspace{3mm}\\
\multicolumn{2}{c}{\includegraphics[width=0.72\columnwidth]{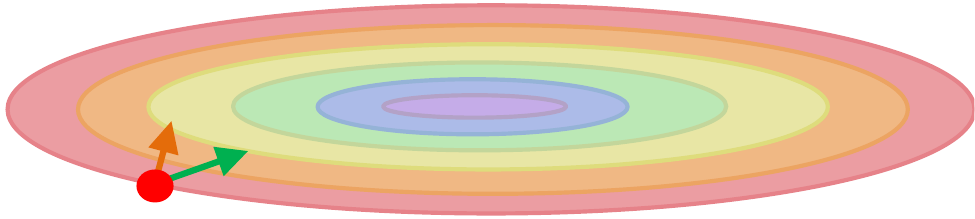}}\\
\multicolumn{2}{c}{(e)}
\end{tabular}
\vspace{-1mm}
}
\caption{Visualization of quadratic models $p$ in~\cref{e:quadpolymodel} for different matrices $\mathbf{A}$ in (a)--(d). (e) a top-down view of (b), illustrating level curves and search directions: steepest descent (orange arrow, orthogonal to the level curves; \cref{e:firstorderopt}) and Newton’s method (green arrow; \cref{e:secondorderopt}) at a hypothetical search point (red circle).
}
\label{f:quadpolymodel}
\end{figure}

\vspace{3mm}
\noindent\textbf{Quadratic polynomial model functions.\;}
To facilitate the subsequent discussion, we consider simple two-dimensional quadratic polynomial model functions (\cref{f:quadpolymodel}):
\begin{align}
\label{e:quadpolymodel}
p(\bm{\theta}) = \frac{1}{2} \bm{\theta}^\top \mathbf{A} \bm{\theta}+\mathbf{b}^\top\bm{\theta}+c,
\end{align}
where $\mathbf{A}$ is a symmetric matrix. For simplicity, we assume $\mathbf{b}=0$ and $c=0$. Since the Hessian of $p$ is constant and equal to $\mathbf{A}$, analyzing its properties becomes straightforward. 

For diagonal matrices $\mathbf{A}$ (\cref{f:quadpolymodel}(a--c)), the diagonal elements $a^{11}$ and $a^{22}$ correspond to the second-order derivatives along each coordinate axis, determining the curvature of $p$. Specifically, when $\mathbf{A}=\mathbf{I}$ (\cref{f:quadpolymodel}(a)), $p$ exhibits \emph{isotropic} curvature, \ie, it has uniform curvature in all directions. In (b), where $a^{11}$ is significantly larger in magnitude than $a^{22}$, $p$ is more strongly curved along the first coordinate axis. A top-down view (\cref{f:quadpolymodel}(e)) reveals an elongation of the corresponding level curves. In (c), where $a^{11}$ is negative, $p$ is concave along the first coordinate but remains convex along the second, making the function neither fully convex nor concave. 

For general non-diagonal matrices (\eg, \cref{f:quadpolymodel}(d)), curvature analysis is facilitated by eigenvalue decomposition:
\begin{align}
\label{e:eigendecomposition}
\mathbf{A} = \lambda_1\mathbf{e}_1\mathbf{e}_1^\top+\lambda_2\mathbf{e}_2\mathbf{e}_2^\top,
\end{align}
where $\lambda_1$ and $\lambda_2$ are the largest and smallest eigenvalues in absolute magnitude, respectively. The eigenvector $\mathbf{e}_1$ aligns with the direction of maximum curvature, while $\mathbf{e}_2$ corresponds to the direction of minimal curvature. In case $(d)$, with $\lambda_1=10$ and $\lambda_2=1$, the function retains the overall shape of (b) but it is elongated along the direction of $\mathbf{e}_2$.

This explains how Newton's method adjusts the search direction using the inverse Hessian $\mathbf{H}^{-1}$:
\begin{align}
\label{e:newtondirection}
\mathbf{H}^{-1}\nabla p(\bm{\theta}) = \frac{1}{\lambda_1}\mathbf{e}_1\left(\mathbf{e}_1^\top\nabla p(\bm{\theta})\right)+\frac{1}{\lambda_2}\mathbf{e}_2\left(\mathbf{e}_2^\top\nabla p(\bm{\theta})\right).
\end{align}
Effectively, Newton's method amplifies movement along low-curvature directions in $\nabla p(\bm{\theta})$, as illustrated in \cref{f:quadpolymodel}(e).

\vspace{3mm}
\noindent\textbf{Hessian estimation.\;}
Inspired by finite-element methods for solving high-order differential equations~\cite{XGZ19} and model-based trust-region methods for gradient-free optimization~\cite{NW06}, we estimate the Hessian $\mathbf{H}$ as the coefficient matrix $\mathbf{A}$ of a quadratic polynomial $p$ (\cref{e:quadpolymodel}) fitted to $f$ at $\bm{\theta}$. The zeroth-order term $c$ and the first-order term $\mathbf{b}$ are set to $f(\bm{\theta})$ and $\widehat{\mathbf{g}}$ (see~\cref{e:ourgradient}), respectively. Determining $\mathbf{A}$ requires at least three input points $\Theta=\{\overline{\bm{\theta}}_i\}$ with corresponding function evaluations, in addition to the two points used for computing $\widehat{\mathbf{g}}$. The process for acquiring $\Theta$ is described shortly.

Given these $f$ evaluations, the Hessian $\mathbf{A}$ is determined using standard regularized least squares approach: 
\begin{align}
\label{e:lsq}
\min_{\mathbf{h}\in \mathbb{R}^3} \sum_{i=1}^{|\Theta|}\left(q(\overline{\bm{\theta}}_i)-\phi(\overline{\bm{\theta}}_i)^\top\mathbf{h}\right)^2,
\end{align}
where $\phi(\bm{\theta})=[\frac{1}{2}\theta^1\theta^1,\theta^1\theta^2,\frac{1}{2}\theta^2\theta^2]^\top$ encodes second-order monomials with $\theta^j$ being the $j$-th element of $\bm{\theta}$, and
\begin{align}
q(\overline{\bm{\theta}}_i)=f(\overline{\bm{\theta}}_i)-\widehat{\mathbf{g}}^\top \overline{\bm{\theta}}_i-f(\bm{\theta}).
\end{align}
The minimizer $\mathbf{h}_*\in \mathbb{R}^3$ of \cref {e:lsq} is obtained as the solution of a $3\times 3$ linear system:
\begin{align}
\bm{\Phi}^\top \bm{\Phi}\mathbf{h}_*&=\bm{\Phi}^\top \mathbf{q}, \text{ where }\\
\bm{\Phi} &= \left[\phi(\overline{\bm{\theta}}_1)^\top, \ldots, \phi(\overline{\bm{\theta}}_{|\Theta|})^\top\right]^\top,\\
\mathbf{q} &= \left[q(\overline{\bm{\theta}}_1),\ldots,q(\overline{\bm{\theta}}_{|\Theta|})\right]^\top.\nonumber
\end{align}
Once the minimizer $\mathbf{h}_*\in \mathbb{R}^3$ of \cref {e:lsq} is obtained, $\mathbf{A}$ is reconstructed by appropriately arranging the elements of $\mathbf{h}_*$.

\vspace{3mm}
\noindent\textbf{Acquiring sample $f$-evaluations $\Theta$.\;}
\label{p:period} A straightforward approach to estimating the Hessian matrix $\mathbf{H}=\mathbf{A}$ is to randomly sample three input parameter combinations $\{\overline{\bm{\theta}}_i\}_{i=1}^3$ near $\bm{\theta}$ and evaluate the function $f$. However, this increases the number of function evaluations per step, potentially slowing convergence in terms of total function queries. To mitigate this, we employ a \emph{caching} mechanism that stores function evaluations for reuse in subsequent optimization steps. For this strategy to be effective, the subspaces $\mathcal{V}$ and $\{\mathcal{W}\}_{j=1}^\frac{m}{2}$ must be retained. 

To maintain this consistency, we implement a \emph{periodic subspace switching} strategy, where $\mathcal{V}$ and $\{\mathcal{W}\}_{j=1}^\frac{m}{2}$ are updated only every $T$-th optimization step. During intermediate steps $k$ (\ie $k$ mod $T \neq 0,1$), function evaluations used for gradient estimation are reused from the two preceding steps:
\begin{align}
\label{e:perturbation}
\Theta^k &= \{\bm{\theta}_1^{k-2},\bm{\theta}_2^{k-2},\bm{\theta}_1^{k-1},\bm{\theta}_2^{k-1}\}.
\end{align}
Since no function evaluations are available for reuse when $k$ mod $T = 0$, we sample three new random points and evaluate $f$ at these locations. 

\begin{algorithm}[t]
\small
\caption{Subspace-based Hessian Algorithm.}
\label{a:mainalgorithm}
\begin{algorithmic}[1]
  \State \textbf{Input:} Initial search point $\mathbf{x}^0$.
  \State \textbf{Parameters:} Step size (schedule) $\eta$, subspace switching period $T$, maximum iterations $K$.
  \State \textbf{Output:} Final solution $\mathbf{x}^K$.
  \For{$k=0$ to $K-1$}
    \If {$k \bmod T = 0$}
        \State \textbf{Select} an intermediate subspace $\mathcal{V}\subset\mathbb{R}^{m}$ from $\mathbb{R}^d$;
        \For{$j=0$ to ${\frac{m}{2}}-1$}
            \State \textbf{Sample} two indices from $\{1,\ldots,m\}$ without\\ \qquad\quad\quad replacement to define subspace $\mathcal{W}_j\subset\mathbb{R}^2$;\hspace{-0.5cm}\Comment{\CComments{\cref{s:randomsubspace}}}
        \State \textbf{Construct} projection operator $P_j$; \Comment{\CComments{\cref{e:projection}}}
        \EndFor
    \EndIf
    \State $\mathbf{v}^k \gets\mathbf{0}$;
     \For{$j =0$ to ${\frac{m}{2}}-1$}
     \State \textbf{Acquire} sample points $\Theta$ and evaluate $f$ on them;\\ 
    \Comment{\CComments{Reuse $f$ values from the previous steps\\
    \quad\quad\quad\quad\quad\quad\quad\quad\quad\: when available; \cref{s:estimgradhess}}}
    \State \textbf{Estimate} the subspace gradient $\widehat{\mathbf{g}}_j$; \Comment{\CComments{\cref{e:ourgradient}}
    }
    \State \textbf{Estimate} the subspace Hessian 
    $\widehat{\mathbf{H}}_j$; \Comment{\CComments{\cref{e:lsq}}}
    \State \textbf{Accumulate:} 
    $\mathbf{v}^k\gets \mathbf{v}^k+P^{-1}(\widehat{\mathbf{H}}_j^{-1}\widehat{\mathbf{g}}_j)$; \Comment{\CComments{\cref{e:ourupdate}}}
    \EndFor
  \State \textbf{Update:} 
  $\mathbf{x}^{k+1}\gets\mathbf{x}^k-\eta \mathbf{v}^k$;
  \EndFor
  \State \Return $\mathbf{x}^k$;
\end{algorithmic}
\end{algorithm}

\vspace{3mm}
\noindent\textbf{Optimization algorithm.\;}
\Cref{a:mainalgorithm} summarizes our main algorithm for optimizing a convex function $f$. For general non-convex functions, the Hessian matrix $\mathbf{H}$ and its estimated counterpart $\widehat{\mathbf{H}}=\mathbf{A}$ are not necessarily PD. Consequently, the resulting search direction $\mathbf{A}^{-1}\widehat{\mathbf{g}}$ may increase $f$ instead of decreasing it. To prevent this, we enforce positive definiteness by modifying the estimated Hessian $\mathbf{A}$. Specifically, we reconstruct $\mathbf{A}$ by taking the absolute values of its eigenvalues (see~\cref{e:eigendecomposition}). 

The impact of this transformation becomes clear when examining the behavior of the quadratic polynomial model function in \cref{f:quadpolymodel}. First, taking the absolute values of the eigenvalues ensures that the modified Hessian $\overline{\mathbf{A}}$ remains PD, thereby avoiding search directions that could lead to an increase in $f$. Second, the efficiency of second-order methods lies in their ability to adjust search directions based on the function's curvature. Since the elongation (curvature) of $f$ is determined by the absolute magnitudes of the Hessian eigenvalues, this transformation preserves the key benefits of second-order updates. 

When $\mathbf{A}$ contains negative eigenvalues, directly applying $\mathbf{A}^{-1}$ could generate search directions that incorrectly move along concave regions (\cref{f:quadpolymodel}(c)). By taking absolute values, we effectively reverse the movement along these eigenvectors, ensuring that the update remains well-posed for optimization (\cref{f:quadpolymodel}(e)). Following \cite{ZHJ24}, we further improve numerical stability through eigenvalue clipping, replacing each eigenvalue $\lambda_i$ with $\max(\lambda_i, \kappa)$, where the fixed positive parameter $\kappa$ is set to 0.1.

\vspace{3mm}
\noindent\textbf{Discussion.\;}
A key limitation of our approach is that each update step can only capture the joint dependence of at most two parameters. This design reflects the trade-off between incorporating second-order information and minimizing function queries in ZO optimization setting. 

In traditional optimization, the primary concerns are computational and memory complexities. Second-order methods generally incur cubic computational costs per iteration due to Hessian inversion, but they often require fewer iterations to converge to a local minimum. This balance frequently results in lower overall computational complexity. However, our preliminary experiments revealed that even for relatively low-dimensional update subspaces (\eg, $m=4$), using a full second-order approach consistently resulted in a substantially higher number of function queries.

In contrast, our two-dimensional subspace approach requires only three additional function evaluations per update step, with the periodic subspace switching strategy further enabling the complete omission of function evaluations at specific steps. This yields substantially faster overall convergence in terms of function queries while still incorporating curvature information.

An alternative approach is to approximate only the diagonal components of the Hessian~\cite{ZDY25}, offering linear scaling with respect to the function dimensionality. This method allows for the joint consideration of all parameters in each update step. However, its primary limitation is that it accurately captures curvature only when the eigenvectors of the Hessian $\mathbf{H}$ are aligned with the coordinate axes, as shown in \cref{f:quadpolymodel}(a–c). Specifically, diagonal approximations struggle with anisotropic curvature in scenarios like \cref{f:quadpolymodel}(d), where the eigenvectors are rotated by 45 degrees relative to the coordinate axes.

Our experiments demonstrate that the subspace-based approach of ZO-SAH provides substantial overall improvements over diagonal Hessian approximations.

\subsection{Convergence analysis}
\label{s:convergence}
We present the main convergence results of our method below. A complete proof is provided in \cref{s:convergence_supp}.

\begin{assumptions}[Function properties]
\label{a:assumptions1}
The objective function $f: \mathbb{R}^d \to \mathbb{R}$ is $\mu$-strongly convex, bounded below, with Lipschitz-continuous gradient (constant $C_{1}$) and Hessian (constant $C_{2}$). 
\end{assumptions}

\begin{assumptions}[Optimization steps]
\label{a:assumptions2}
Following the setup of \cite{WQY24}, we assume that the \emph{Armijo backtracking line search}~\cite{NW06} with a minimum step size is applied. Specifically, for a search direction $\mathbf{v}^{k}$, there exist bounded constants $c_{1}$, $\rho^{\min}$, and $\rho^{k}$ satisfying $\rho^{\min}\leq \rho^{k}$, such that:
\begin{align}
\label{e:armijo}
f(\mathbf{x}^{k+1}) \leq f(\mathbf{x}^{k})-c_{1} \rho^{k} \|\mathbf{v}^{k}\|^{2}.
\end{align}
We further bound the Hessian perturbation in the subspace $\mathcal{W}$ by $l$ (\ie the distance between $\bm{\theta}^k$ and elements of $\Theta^k$ is at most $l$; see~\cref{e:perturbation}). The gradient perturbation $\epsilon^k$ (\cref{e:ourgradient}) is similarly bounded by $\frac{\sqrt{2}}{dC_1} \|\nabla f(\mathbf{x}^{k})\|$. Finally, the smallest eigenvalue of $\bm{\Phi}^\top \bm{\Phi}$ is bounded below by $\gamma$.
\end{assumptions}
These assumptions, except for the last one which is specific to ZO-SAH, are commonly adopted in ZO optimization studies, and further discussions can be found in~\cite{KMR23,WQY24}.  See \Cref{s:boundhessianestim} for a discussion on $\gamma$.

\begin{proposition}\label{p:mainresult}
Under these conditions, starting from the initial search point $\mathbf{x}^0$ (\cref{e:ourupdate}), 
the sequence $(\mathbf{x}^k)$ generated by our algorithm satisfies that for all $k \geq 1$:
\begin{align}
\label{e:convergence}
\mathbb{E}_{\mathbf{P} \sim \mathcal{D}}&[f(\mathbf{x}^k) - f(\mathbf{x}^*)]
\le \frac{C_1}{2}(1-\nu)^{k}\|\Delta_{0}\|^{2},\\
\nu &= c_1 \rho^{\min} \,\frac{\mu m}{d^{2}\bigl(C_1 + E + \kappa\bigr)^{2}},
\end{align}
where $\mathbf{x}^*$ is the minimizer of $f$, $\mu$ is the strong convexity parameter of $f$, $m$ is the subspace dimension $\mathcal{V}$, 
$\Delta_0$ is the initial optimization gap ($\Delta_0=\|\mathbf{x}^{0}-\mathbf{x}^{*}\|$), and $\mathcal{D}$ is the 
distribution of projection matrices for random coordinate selection (\cref{e:projection};~\cite{GKL19}). 
The term $E$ is defined as:
\begin{align}
E:=
      \sqrt{6}\frac{s}{\gamma}\,
      \left(
         \frac{C_{1}}{d}\Delta_{0}l^{3} 
         \;+\;
         \frac{C_{2}}{6}l^{5}
      \right).
\end{align}
where $s$ is the number of perturbations, fixed at 4 ($s=|\Theta^k|$; see~\cref{e:perturbation}).
\end{proposition}

\begin{sketchofproof}
We first establish an upper bound on the difference between the estimated gradient $\widehat{\mathbf{g}}$ (\cref{e:ourgradient}) in the subspace $\mathcal{W}$ and the exact gradient $\nabla f$ projected onto $\mathcal{W}$:
\begin{align}
\label{e:gradientbound}
  \|\widehat{\mathbf{g}} - \mathbf{P}\nabla f\| \leq \frac{C_{1}\epsilon^{k}}{\sqrt{2}}.
\end{align}
Next, using \cref{e:gradientbound} and the Lipschitz continuity of the gradient and Hessian of $f$, we derive a bound on the spectral norm of the difference between the estimated subspace Hessian $\widehat{\mathbf{H}}$ and the exact Hessian $\nabla^{2} f(\mathbf{x})$ projected onto $\mathcal{W}$: 
\begin{align}
\label{e:Hessianbound}
\|\widehat{\mathbf{H}}-\mathbf{P}\nabla^{2} f(\mathbf{x})\,\mathbf{P}\| \le E.
\end{align}
Applying \cref{e:gradientbound} and \cref{e:Hessianbound} to $\mathbf{x}^k$  and combining these bounds, we obtain the following  lower bound for the update:
\begin{align}
\label{e:updatebound}
\frac{m}{2d^2(C_1+E+\kappa)^{2}} \|\nabla f(\mathbf{x}^{k})\|^{2}
\leq
\|\mathbf{v}^{k}\|^{2}.
\end{align}
Our Hessian estimate $\widehat{\mathbf{H}}$ is PD, ensuring that $-\mathbf{v}^k$ is a descent direction. Finally, substituting \cref{e:updatebound} into \cref{e:armijo}, applying the Polyak--{\L}ojasiewicz inequality~\cite{KNS16}, and telescopically summing from $0$ to $k$ yields \cref{e:convergence}.
\vspace{-2mm}
\end{sketchofproof}

\begin{figure}[bt]
    \centering
    \includegraphics[width=0.95\linewidth]{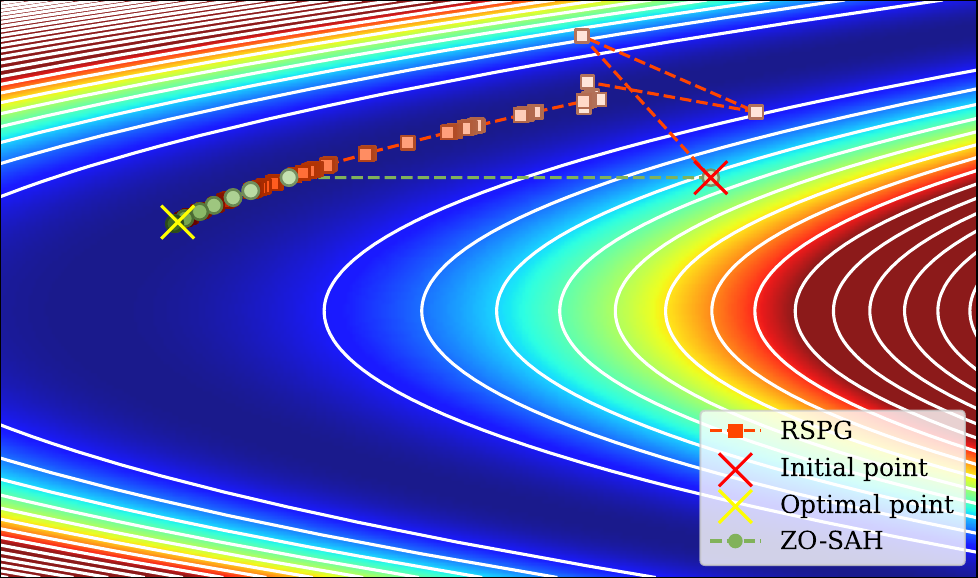}
    \caption{Optimization trajectories of \textbf{\texttt{ZO-SAH}} and \textbf{\texttt{RSPG}} on the toy example in \cref{e:test_function}. Exploiting second-order information, \textbf{\texttt{ZO-SAH}} achieves significantly faster convergence (201 function calls) than \textbf{\texttt{RSPG}} (7,921 function evaluations).
    }
    \label{f:testfunction}
\end{figure}

\section{Experiments}
We evaluated the effectiveness of the proposed \textbf{\texttt{ZO-SAH}} algorithm on eight benchmark datasets, including six linear classification tasks using logistic regression and two image classification benchmarks using deep neural networks with two different architectures. We also present a simple two-dimensional optimization problem to illustrate how exploiting second-order curvature information improves performance over conventional zeroth-order approaches.\footnote{The code for our algorithm will be released online upon acceptance.}

\begin{figure*}[hbt!]
    \centering
    \begin{minipage}{\linewidth}
        \begin{subfigure}{0.33\linewidth}
            \centering
            \includegraphics[width=\linewidth]{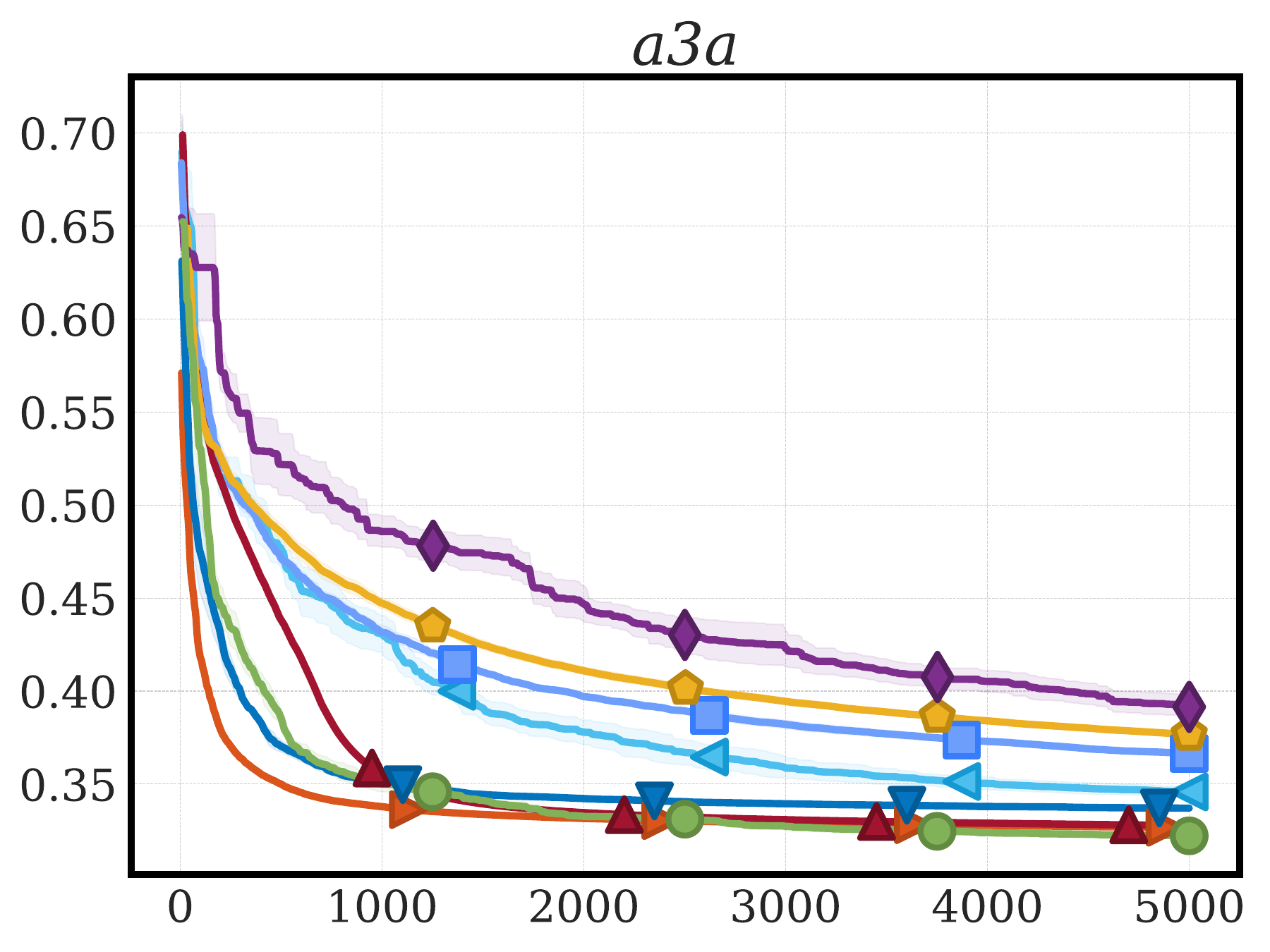}
            \label{sf:a3a}
        \end{subfigure}
        \begin{subfigure}{0.33\linewidth}
            \centering
            \includegraphics[width=\linewidth]{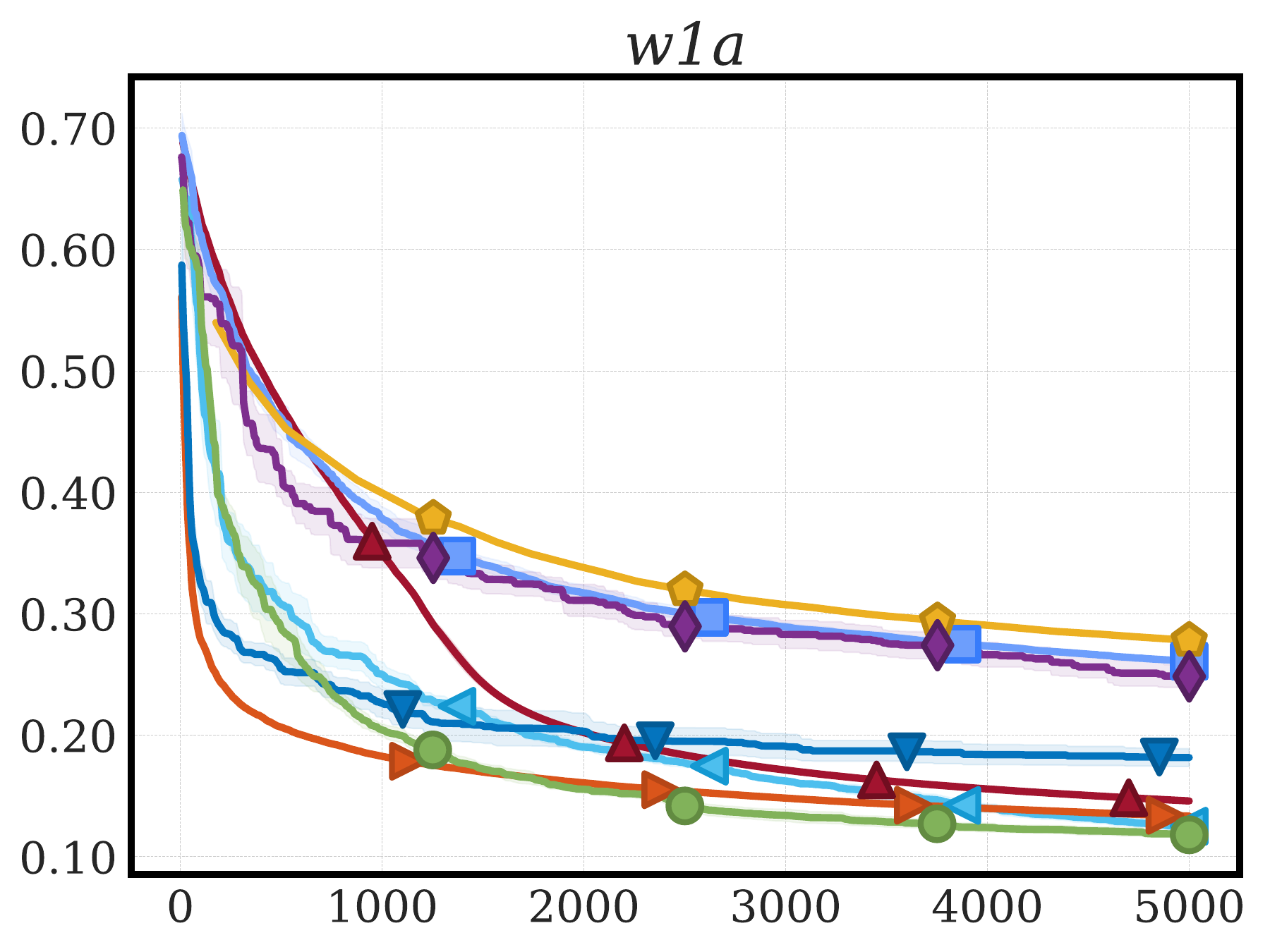}
            \label{sf:a5a}
        \end{subfigure}
        \begin{subfigure}{0.33\linewidth}
            \centering
            \includegraphics[width=\linewidth]{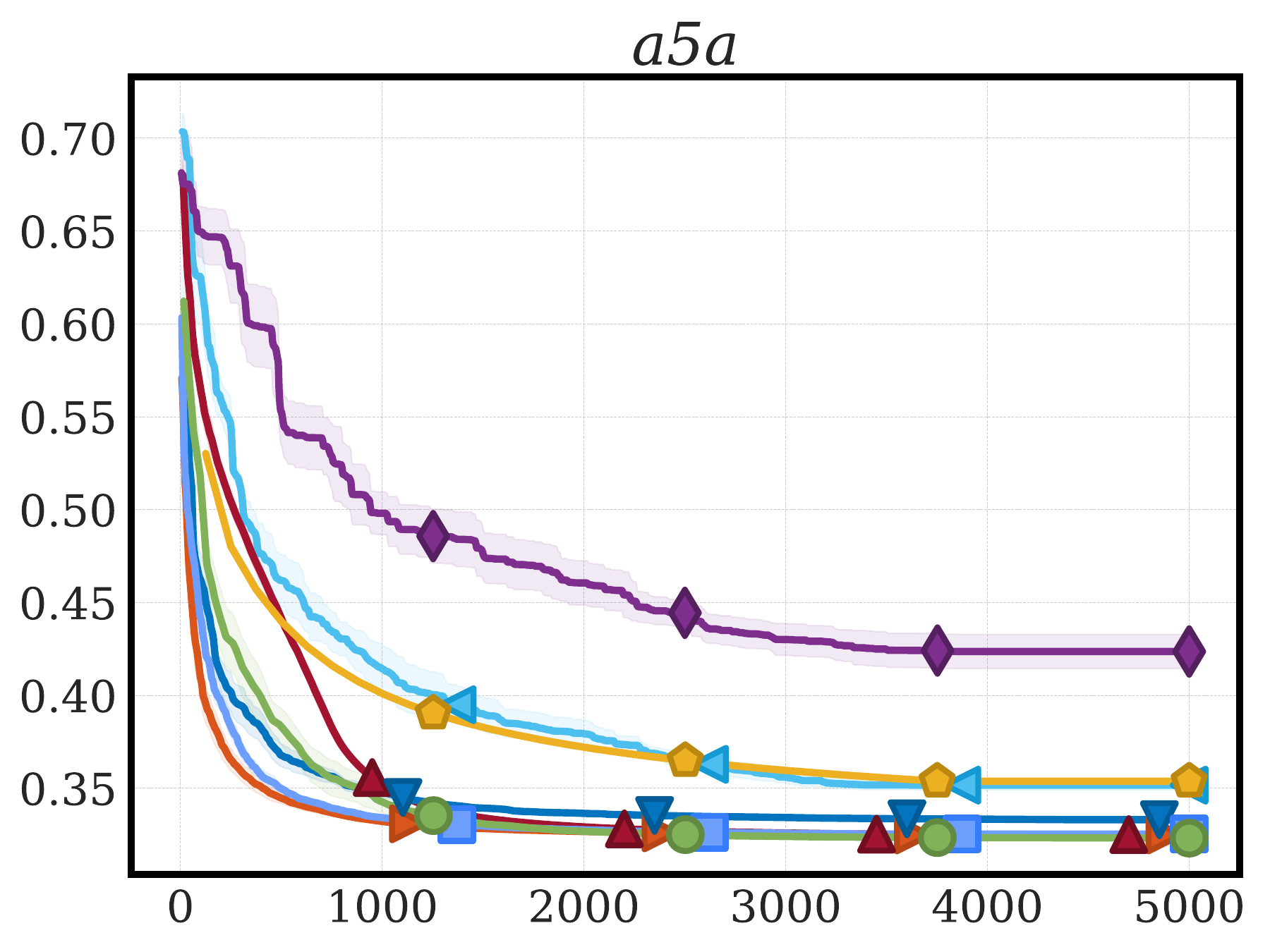}
            \label{sf:w1a}
        \end{subfigure}  
    \end{minipage}
    \vspace{-10pt}

    \begin{minipage}{\linewidth}
        \begin{subfigure}{0.33\linewidth}
            \centering
            \includegraphics[width=\linewidth]{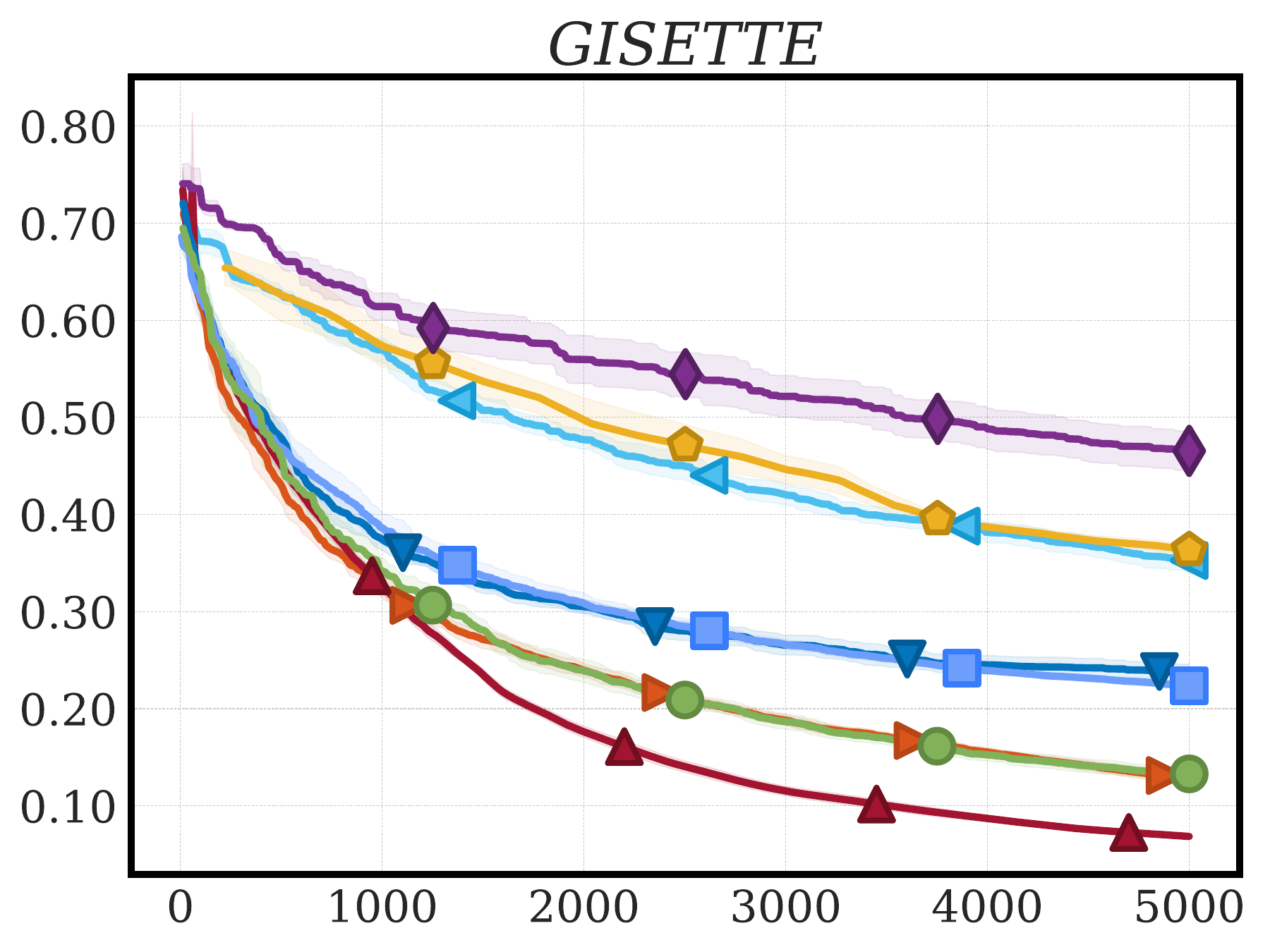}
            \label{sf:gis}
        \end{subfigure}
        \begin{subfigure}{0.33\linewidth}
            \centering
            \includegraphics[width=\linewidth]{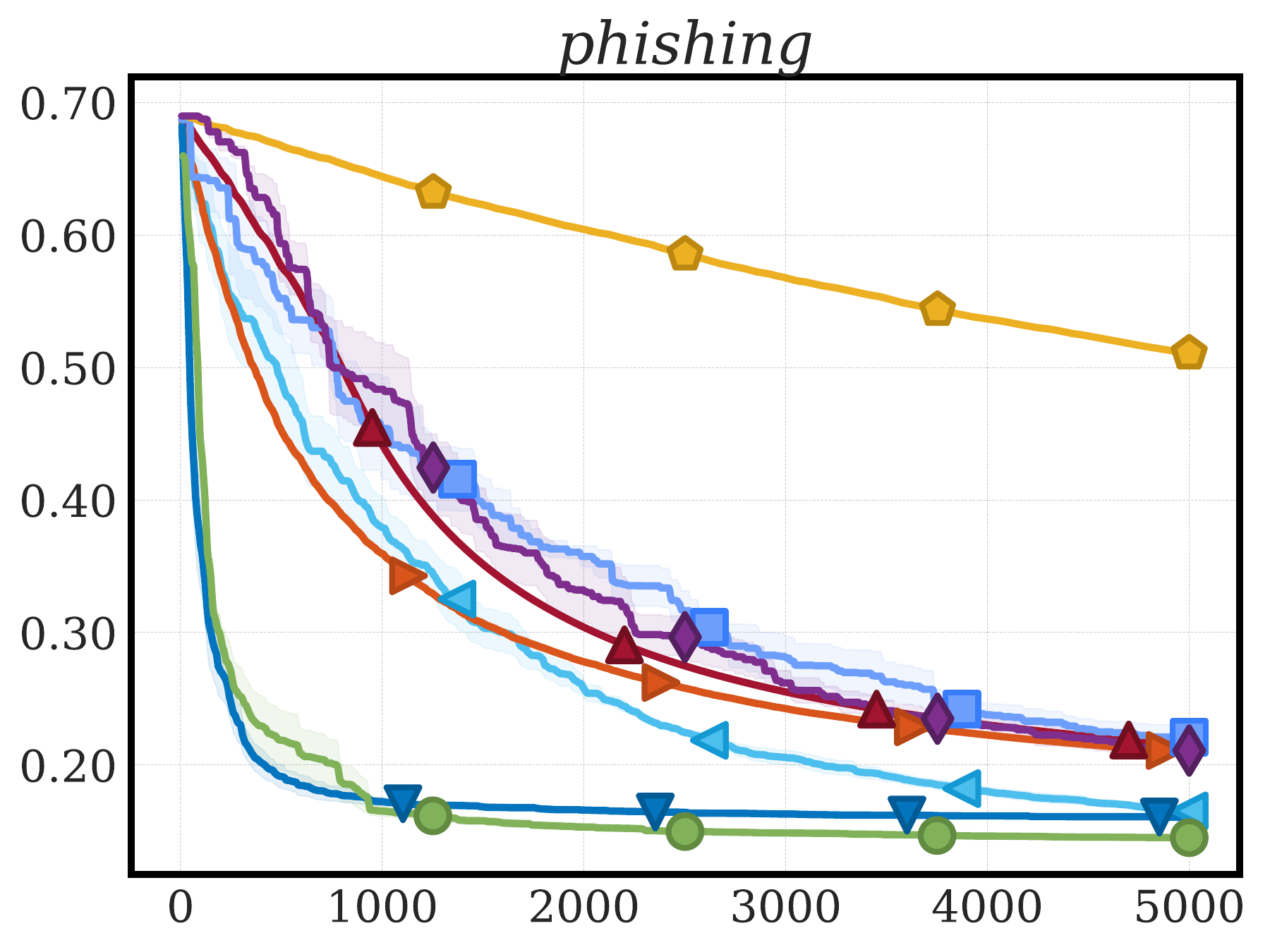}
            \label{sf:phi}
        \end{subfigure}
        \begin{subfigure}{0.33\linewidth}
            \centering
            \includegraphics[width=\linewidth]{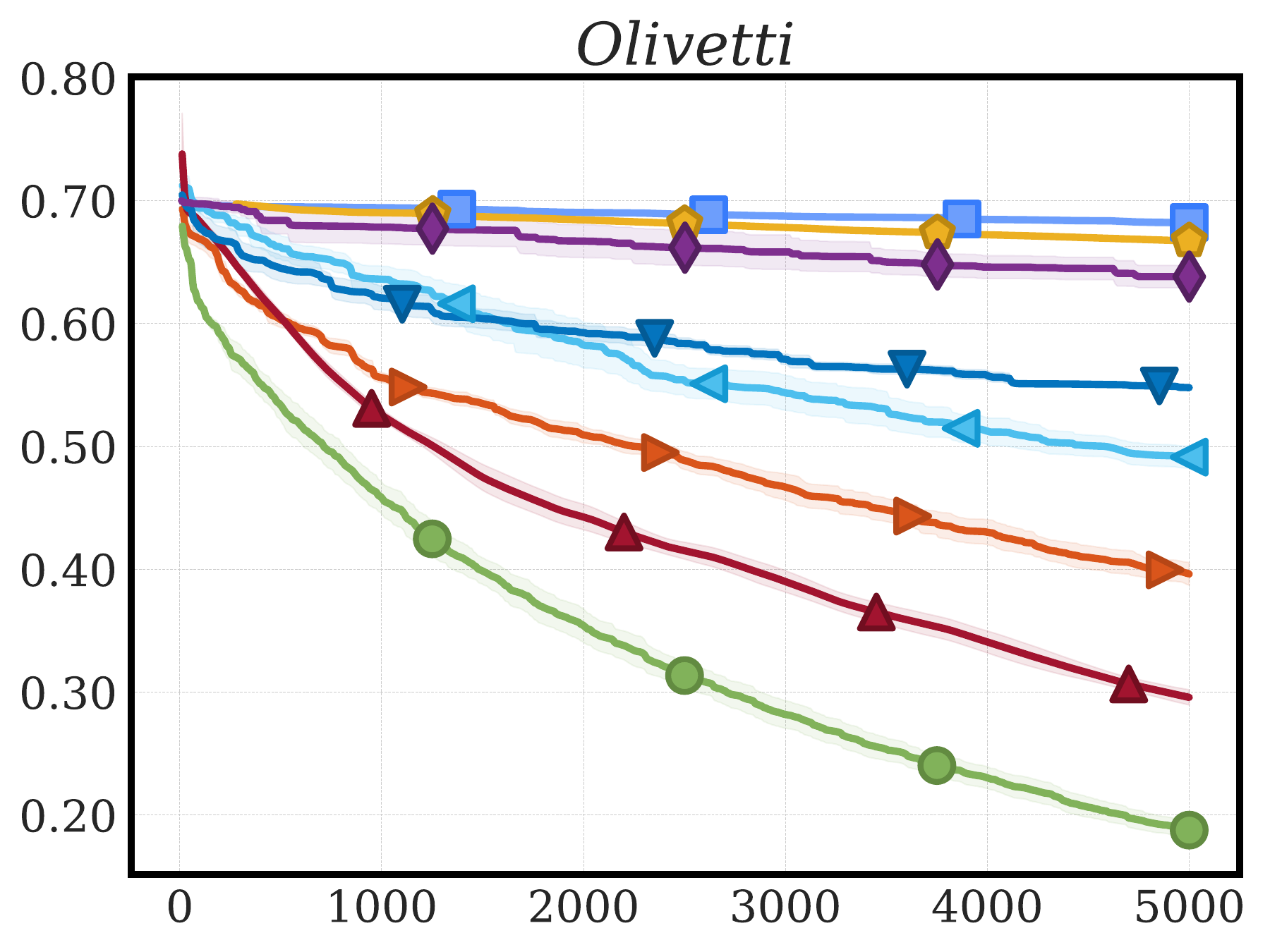}
            \label{sf:oli}
        \end{subfigure}  
    \end{minipage}
    \vspace{-10pt}
    
    \begin{minipage}{\linewidth}
        \begin{subfigure}{0.33\linewidth}
            \centering
            \includegraphics[width=\linewidth]{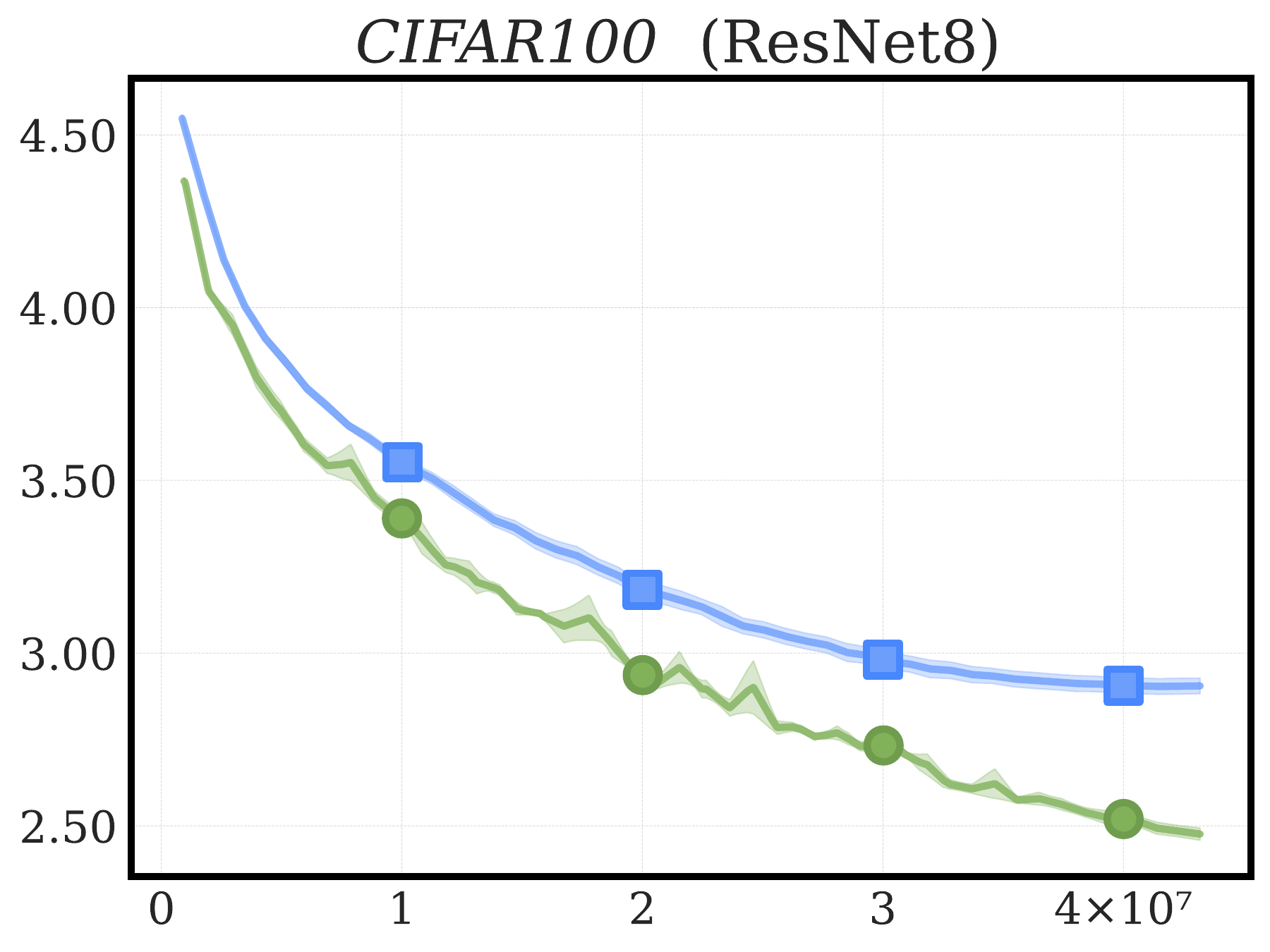}
            \label{sf:resnet20cifar10}
        \end{subfigure}
        \begin{subfigure}{0.33\linewidth}
            \centering
            \includegraphics[width=\linewidth]{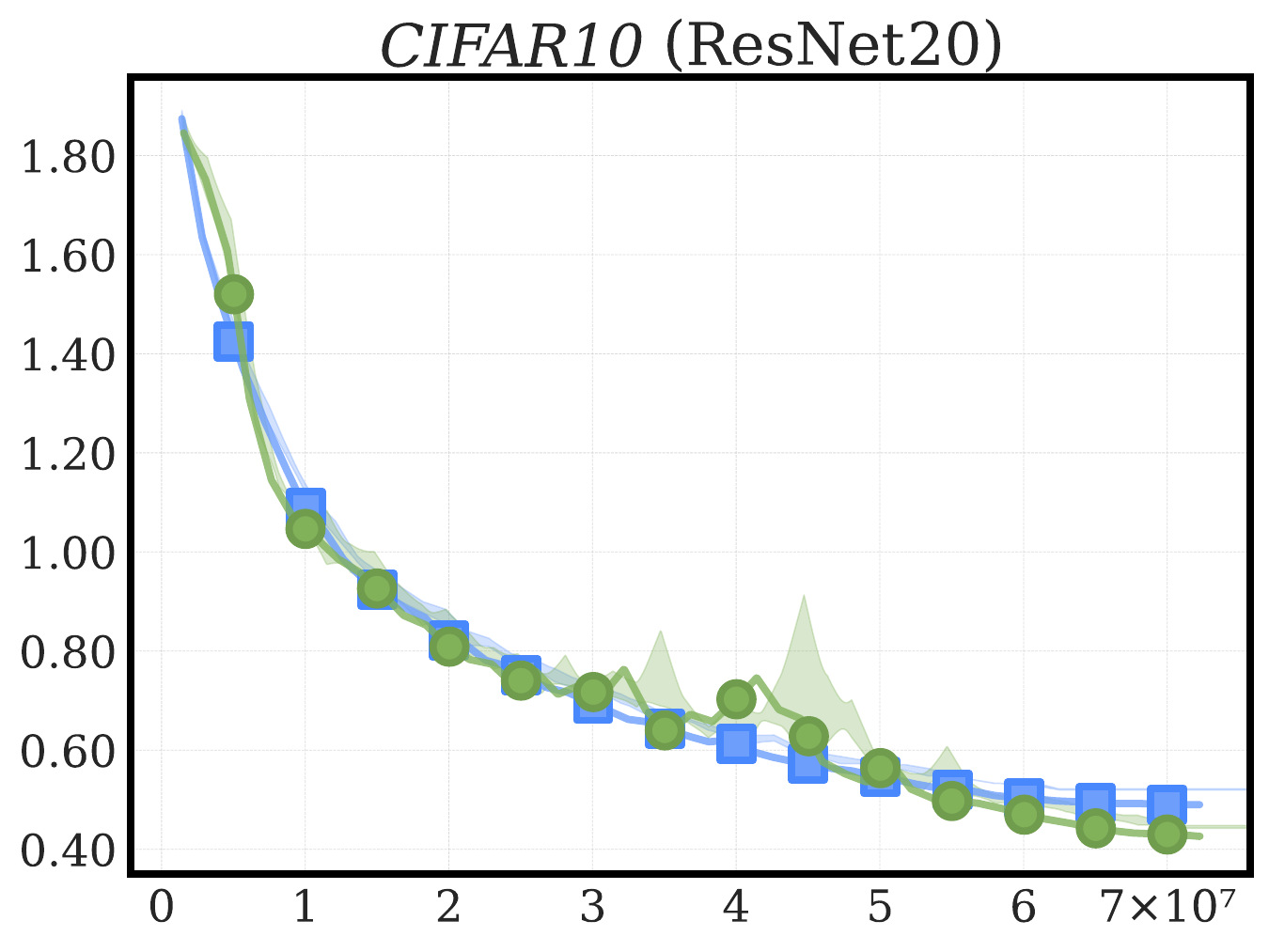}
            \label{sf:resnet8cifar10}
        \end{subfigure}
        \begin{subfigure}{0.33\linewidth}
            \centering
            \includegraphics[width=\linewidth]{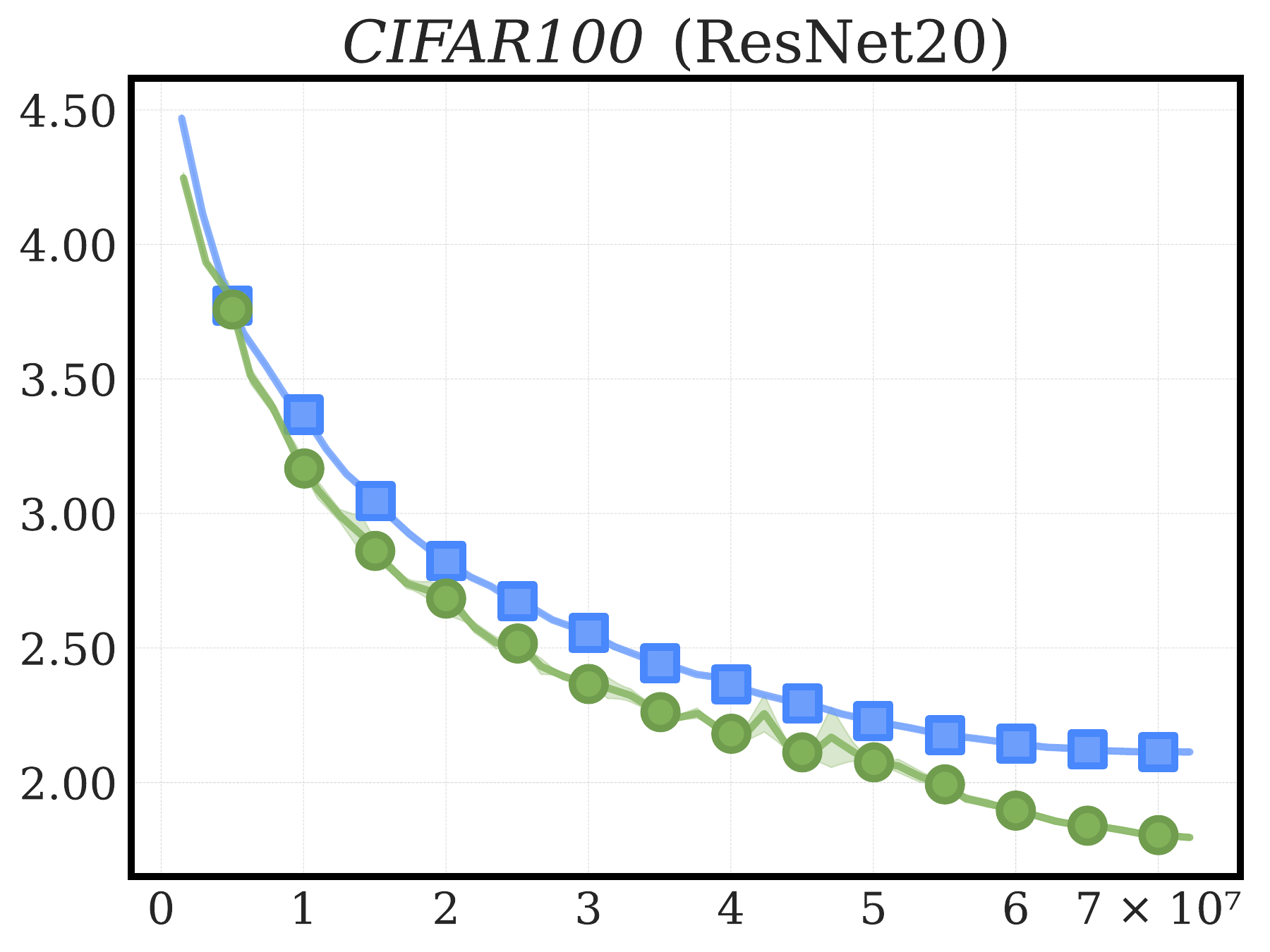}
            \label{sf:resnet20cifar100}
        \end{subfigure}  
    \end{minipage}
    \vspace{-10pt}

    \begin{subfigure}{\linewidth}
        \raggedleft
        \includegraphics[width=0.971\linewidth]{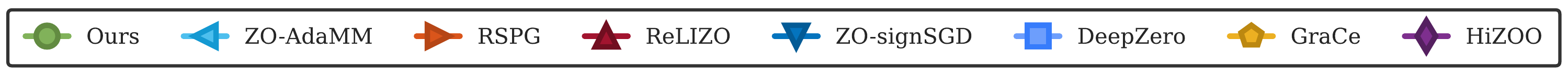}
    \end{subfigure}
    \caption{Progression of average objective function values (y-axis) across eight benchmark datasets. Since each algorithm step requires a different number of function evaluations, the x-axis is aligned to show the cumulative number of function evaluations. The first two rows correspond to logistic regression tasks, while the last row focuses on neural network training. Shaded regions represent twice the standard deviation.}
    \label{f:mainresults}
\end{figure*}

\subsection{Illustrative example}
We consider the standard Rosenbrock function, known for its highly anisotropic landscape:
\begin{align}
\label{e:test_function}
f(x,y)=(x-1)^{2}+100(y-x^2)^2.
\end{align}
\Cref{f:testfunction} shows the optimization trajectories of \textbf{\texttt{ZO-SAH}} and the standard \emph{randomized stochastic projected gradient} (\textbf{\texttt{RSPG}})~\cite{GLZ16}. As a first-order approximation, \textbf{\texttt{RSPG}} initially exhibits typical first-order behavior: its search direction (opposite to the gradient) remains orthogonal to the level curves, leading to oscillatory movement. While \textbf{\texttt{RSPG}} eventually aligns with the valley's curvature, convergence remains slow due to persistent small-step oscillations.

In contrast, \textbf{\texttt{ZO-SAH}} effectively incorporates curvature information to generate more informed search directions from the start. This results in smoother trajectories, avoiding oscillatory behavior and enabling the algorithm to reach the optimal solution in significantly fewer steps.

\subsection{Logistic regression}
\label{s:logistic_result}
\paragraph{Setting.}
Given input vectors and labels $\{(\mathbf{z}_i,y_i)\}_{i=1}^N$, we minimize the logistic regression cost function:
\begin{align}
f(\mathbf{x}) := \frac{1}{N} \sum_{i=1}^{N} \ln \left( 1 + \exp\left(-y_{i} \mathbf{z}_{i}^{\top} \mathbf{x}\right) \right),
\end{align}
applied to six benchmark datasets: \emph{a3a}~\cite{Pla99}, \emph{a5a}~\cite{Pla99}, \emph{w1a}~\cite{Pla99}, \emph{GISETTE}~\cite{GGB04}, Olivetti Faces (\emph{Olivetti})~\cite{SH94}, and \emph{phishing} dataset~\cite{ml12}. Detailed dataset information is provided in~\cref{sss:dataset}. 

To benchmark our approach, we evaluated several existing ZO optimization methods: \textbf{\texttt{RSPG}}, \textbf{\texttt{ZO-signSGD}}~\cite{LCC19}, \textbf{\texttt{ZO-AdaMM}}~\cite{CLX19}, \textbf{\texttt{ReLIZO}}~\cite{WQY24}, \textbf{\texttt{GraCe}}~\cite{QT24}, \textbf{\texttt{HiZOO}}~\cite{ZDY25}, and \textbf{\texttt{DeepZero}}~\cite{CZJ24}. Additional information on these methods is provided in~\cref{sss:baseline}. All experiments were repeated ten times, and the convergence results were averaged.

\vspace{3mm}
\noindent\textbf{Results.\;}
The first two rows of \cref{f:mainresults} show the convergence results for up to 5,000 function evaluations, with extended results provided in \cref{sss:extended}. The performance of different algorithms varies widely across datasets, with no single method consistently leading in all scenarios. While \textbf{\texttt{ReLIZO}} achieves the best convergence on the \emph{GISETTE} dataset, it performs significantly worse than the baseline \textbf{\texttt{RSPG}} on the \emph{Olivetti} and \emph{phishing} datasets.

Apart from the \emph{phishing} dataset, \textbf{\texttt{RSPG}} consistently demonstrates strong performance in terms of average final function values. It also achieves rapid early-stage convergence on \emph{a3a}, \emph{a5a}, and \emph{w1a}. The limited performance gains of \textbf{\texttt{HiZOO}} highlight the challenge of effectively balancing second-order information with function evaluation efficiency. Our algorithm \textbf{\texttt{ZO-SAH}} exhibited robust performance across all datasets, achieving the second-best final average function value on \emph{GISETTE}, following \textbf{\texttt{ReLIZO}}, and outperforming all other methods on the remaining datasets.

The performance on \emph{GISETTE} also highlights a limitation of \textbf{\texttt{ZO-SAH}}. For isotropic functions, second-order methods, including \textbf{\texttt{ZO-SAH}}, may not offer substantial advantages over first-order (or zeroth-order) methods. Analyzing the Hessian condition numbers (\ie the ratio of the largest to smallest eigenvalues) along the optimization trajectories, we observed an average of 3.93 on \emph{GISETTE}, indicating a relatively isotropic function, while \emph{Olivetti} exhibited a more anisotropic landscape with an average condition number of 15.23. Despite this limitation, \textbf{\texttt{ZO-SAH}} maintained a performance on par with \textbf{\texttt{RSPG}}, demonstrating robustness through efficient function evaluation reuse, even in scenarios where curvature information provide limited benefits.

\subsection{Neural networks}
Following the experimental setup of \cite{CZJ24}, we performed image classification experiments using ResNet8 and ResNet20 architectures~\cite{HZR15} on the \emph{CIFAR10} and \emph{CIFAR100} datasets~\cite{KNH09}. We compared \textbf{\texttt{ZO-SAH}} with \textbf{\texttt{DeepZero}}, which, to the best of our knowledge, is the only known ZO optimization method demonstrated to effectively train deep neural networks. Additional training details are provided in \cref{sss:implinfo}.

\begin{figure}[hbt!]
    \centering
    \begin{subfigure}{0.67\linewidth}
       \centering
       \includegraphics[width=\textwidth]{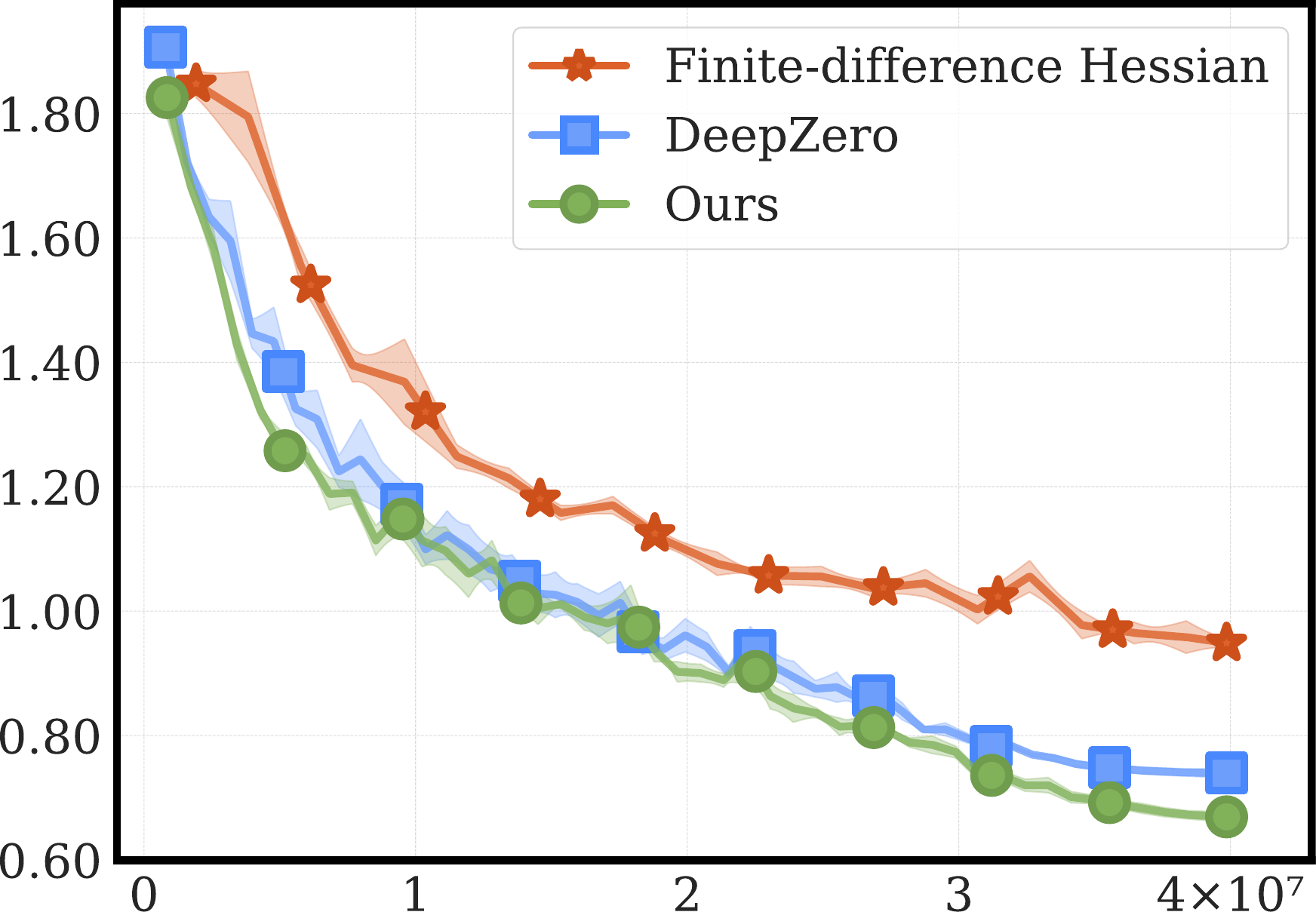}
       \label{sf:res8_10}
       \vspace{-10pt}
   \end{subfigure}
    \caption{Performance of our algorithm and finite-difference Hessian estimation on \emph{CIFAR10} with ResNet8.}
    \label{f:ablationreuse}
\end{figure}

The last row of \cref{f:mainresults} and \cref{f:ablationreuse} presents the results. \textbf{\texttt{ZO-SAH}} consistently demonstrates faster convergence than \textbf{\texttt{DeepZero}}, achieving more rapid loss reductions with fewer function queries on average. Specifically, when training ResNet8 on \emph{CIFAR10} and ResNet20 on \emph{CIFAR100}, \textbf{\texttt{ZO-SAH}} reaches the final loss level of \textbf{\texttt{DeepZero}} while requiring approximately 50\% and 29\% fewer function queries, respectively. Moreover, \cref{t:cifar} shows that \textbf{\texttt{ZO-SAH}} also achieves higher test accuracy, demonstrating that incorporating second-order information does not adversely affect generalization performance~\cite{ND21}.

\begin{table}[h]
    \centering
    \resizebox{\columnwidth}{!}{
    \setlength{\tabcolsep}{2pt}
    \small
    \begin{tabular}{cccccc}
        \toprule
        \multirow{2}{*}{Model} & \multirow{2}{*}{Dataset} & \multicolumn{2}{c}{\textbf{\texttt{ZO-SAH}}} & \multicolumn{2}{c}{\textbf{\texttt{DeepZero}}} \\
        \cmidrule(r){3-6}
         & & Train & Test & Train & Test \\
        \midrule
        \multirow{2}{*}{ResNet8} & \emph{CIFAR10}  & 76.9$\pm$0.8  & 76.3$\pm$1.3  & 74.3$\pm$0.5  & 73.4$\pm$0.7  \\
        & \emph{CIFAR100}  & 36.0$\pm$0.4  &  35.2$\pm$0.3  & 27.5$\pm$0.3  & 26.7$\pm$0.4 \\
        \midrule
        \multirow{2}{*}{ResNet20} & \emph{CIFAR10}  & 85.3$\pm$0.1 & 83.7$\pm$0.1 & 83.1$\pm$0.1  & 81.6$\pm$0.1  \\
        & \emph{CIFAR100} & 50.8$\pm$0.6  & 48.8$\pm$0.7  & 43.4$\pm$0.5  & 42.4$\pm$0.7  \\
        \bottomrule
    \end{tabular}}
    \caption{Training and testing accuracies (\%) of neural networks optimized with \textbf{\texttt{ZO-SAH}} and \textbf{\texttt{DeepZero}}.}
    \label{t:cifar}
\end{table}

\vspace{4mm}
\noindent\textbf{Comparison with finite difference Hessian estimates.\;}
Our method estimates the Hessian of $f$ in the subspace $\mathcal{Z}$ by fitting a quadratic polynomial, facilitating the reuse of function evaluations from previous iterations. We compared this with the traditional coordinate-wise finite difference method, which requires three function evaluations per iteration. 

Preliminary experiments showed that both approaches exhibit similar performance in terms of the number of gradient descent steps. However, when considering the total number of function evaluations, the finite difference method performed significantly worse than both \textbf{\texttt{DeepZero}} and \textbf{\texttt{ZO-SAH}}, demonstrating the effectiveness of our final design in reducing function query costs~(\cref{f:ablationreuse}). 

Additional ablation results are provided in~\cref{sss:ablation}.

\section{Conclusion}
In this paper, we presented the subspace-based approximate Hessian (ZO-SAH) algorithm, a new approach for zeroth-order optimization that effectively incorporates second-order curvature information while maintaining a low function evaluation cost. ZO-SAH operates by projecting the optimization problem into randomly selected two-dimensional subspaces and estimating the Hessian through quadratic polynomial fitting. This design not only captures valuable curvature information but also enables the reuse of function evaluations across iterations, significantly enhancing function query efficiency. Experiments on eight benchmark datasets demonstrated that ZO-SAH achieves faster convergence and competitive test accuracy across diverse optimization scenarios.

{
    \small
    \bibliographystyle{ieeenat_fullname}
    \bibliography{zosah}
}

\newpage
\onecolumn

\appendix

\begin{center}
\textbf{\LARGE\bfseries Appendix}
\end{center}

\noindent In this appendix, we provide
\begin{enumerate}
\item the full proof of our main convergence result (\cref{p:mainresult}; \cref{s:convergence_supp});
\item additional information, including preliminary details on zeroth-order (ZO) optimization and a detailed description of our experimental settings (\cref{s:expdetails}); and
\item extended experimental results (\cref{s:extendedresults}).
\end{enumerate}
For completeness, some content from \cref{s:convergence} is reproduced here.

\section{Convergence analysis}
\label{s:convergence_supp}
This section presents a complete proof of Proposition~\ref{p:mainresult}. We begin by introducing assumptions commonly used in zeroth-order optimization~\cite{KMR23, WQY24}.

\begin{assumptions}[Function properties]
\label{a:assumptions1_supp}

The objective function $f: \mathbb{R}^d \to \mathbb{R}$ is $\mu$-strongly convex, bounded below, with Lipschitz-continuous gradient (constant $C_{1}$) and Hessian (constant $C_{2}$). 
\end{assumptions}

\begin{assumptions}[Optimization steps]
\label{a:assumptions2_supp}
Following the setup of \cite{WQY24}, we assume that the \emph{Armijo backtracking line search}~\cite{NW06} with a minimum step size is applied. Specifically, for a search direction $\mathbf{v}^{k}$, there exist bounded constants $c_{1}$, $\rho^{\min}$, and $\rho^{k}$ satisfying $\rho^{\min}\leq \rho^{k}$, such that:
\begin{align}
\label{e:armijo_supp}
f(\mathbf{x}^{k+1}) \leq f(\mathbf{x}^{k})-c_{1} \rho^{k} \|\mathbf{v}^{k}\|^{2}.
\end{align}
We further bound the Hessian perturbation in the subspace $\mathcal{W}$ by $l$ (\ie the distance between $\bm{\theta}^k$ and elements of $\Theta^k$ is at most $l$). The gradient perturbation $\epsilon^k$ is similarly bounded by $\frac{\sqrt{2}}{dC_1} \|\nabla f(\mathbf{x}^{k})\|$. Finally, the smallest eigenvalue of $\bm{\Phi}^\top \bm{\Phi}$ is bounded below by $\gamma$.
\end{assumptions}
\noindent Under Assumptions~\ref{a:assumptions1_supp} and \ref{a:assumptions2_supp}, we first bound the deviation between the true analytical gradient and Hessian and their respective estimates computed by our algorithm.

\begin{proposition}\label{p:mainresult_supp}
Under these conditions, starting from the initial search point $\mathbf{x}^0$, 
the sequence $(\mathbf{x}^k)$ generated by our algorithm satisfies that for all $k \geq 1$:
\begin{align}
\label{e:convergence_supp}
\mathbb{E}_{\mathbf{P} \sim \mathcal{D}}&[f(\mathbf{x}^k) - f(\mathbf{x}^*)]
\le \frac{C_1}{2}(1-\nu)^{k}\|\Delta_{0}\|^{2},\\
\nu &= c_1 \rho^{\min} \,\frac{2\mu m}{d^{2}\bigl(C_1 + E + \kappa\bigr)^{2}},
\end{align}
where $\mathbf{x}^*$ is the minimizer of $f$, $\mu$ is the strong convexity parameter of $f$, $m$ is the subspace dimension $\mathcal{V}$, 
$\Delta_0$ is the initial optimization gap ($\Delta_0=\|\mathbf{x}^{0}-\mathbf{x}^{*}\|$), and $\mathcal{D}$ is the 
distribution of projection matrices for random coordinate selection;~\cite{GKL19}). 
The term $E$ is defined as:
\begin{align}
E &=\sqrt{6}\,\frac{s}{\gamma}\left(
     \frac{C_{1}}{d}\,\Delta_0\,l^{3} 
     + \frac{C_{2}\,l^{5}}{6}\right),
\end{align}
where 
$s$ is the number of perturbations, fixed at 4 ($s=|\Theta^k|$).
\end{proposition}

\subsection{Bounds on the gradient and Hessian estimates}
\label{s:boundhessianestim}
\begin{corollary}
Let $\widehat{\mathbf{g}}$ be the gradient estimate in a given 2-dimensional subspace $\mathcal{W}$ of $\mathbb{R}^d$, and let $\mathbf{P} \nabla f(\mathbf{x}^{k})$ denote the analytical gradient projected onto $\mathcal{W}$ via the projection matrix $\mathbf{P}\in \mathbb{R}^{2\times d}$. Then, the difference between them is bounded by  
\begin{align}
\|\widehat{\mathbf{g}} - \mathbf{P} \nabla f(\mathbf{x}^{k})\| \leq \frac{C_1}{\sqrt{2}}\epsilon^{k}.
\end{align}
\label{e:corollary1}
\end{corollary}
\begin{proof}
Let $\mathbf{p}_i$ be the $i$-th row of the matrix $\mathbf{P}$, converted into a column vector. Applying the first-order Taylor expansion with the error term in Lagrange form along the direction $\mathbf{p}_i$ at $f(\mathbf{x}^{k})$, we obtain
\begin{align}
  f(\mathbf{x}^{k} + \epsilon^{k}\,\mathbf{p}_i) 
  = f(\mathbf{x}^{k}) + \epsilon^{k}\,\nabla f(\mathbf{x}^{k})^\top \mathbf{p}_i + \frac{({\epsilon^{k}})^2}{2}\,\mathbf{p}_{i}^\top  
  H f(\mathbf{x}^{k} + \epsilon^{k}\alpha\,\mathbf{p}_{i}) \mathbf{p}_i,
\end{align}
for $\alpha \in [0,1]$. Rearranging the zeroth- and first-order terms and taking absolute values, we get
\begin{align}
\left|\frac{f(\mathbf{x}^{k}+\epsilon^{k}\,\mathbf{p}_i) - f(\mathbf{x}^k)}{\epsilon^{k}} - \nabla f(\mathbf{x}^k)^{\top} \mathbf{p}_i \right| &= \left|\frac{\epsilon^{k}}{2}\,\mathbf{p}_i^{\top} 
H f\Bigl(\mathbf{x}^{k}+\epsilon^{k}\alpha\,\mathbf{p}_i\Bigr) \mathbf{p}_i\right| \leq \frac{\epsilon^{k}}{2}C_{1}.
\end{align}
\noindent Since the matrix $\mathbf{P}$ consists of two rows, it follows from~\cite{ESX22} that 
\begin{align}
\label{e:gradientbound_supp}
    \left\|\widehat{\mathbf{g}} - \mathbf{P}\nabla f(\mathbf{x}^k)\right\| \leq \frac{\epsilon^{k}}{\sqrt{2}}C_1.
\end{align}
\end{proof}
\begin{corollary}
Let $\widehat{\mathbf{H}}$ be the Hessian estimated in $\mathcal{W}$ using the least squares method (see~\cref{s:estimgradhess}). Then, the spectral norm of the difference between the analytical subspace Hessian, $\mathbf{H}=\mathbf{P}\nabla^{2} f(\mathbf{x}^{k})\mathbf{P}^{\top}$, and $\widehat{\mathbf{H}}$ is bounded by  
\begin{align}
\|\widehat{\mathbf{H}} - \mathbf{H}\|_{2} \leq \sqrt{6}\frac{s}{\gamma}\left(\frac{C_1}{\sqrt{2}}\epsilon^{k}l^{3} + \frac{C_2}{6}l^5\right).
\label{corollary2}
\end{align}
\end{corollary}
\begin{proof}
Let $\bm{\theta}^k$ be the projection of $\mathbf{x}^k$ onto $\mathcal{W}$. For simplicity of exposition, we reparameterize the coordinate variables $\bm{\theta}$ such that $\bm{\theta}^k$ is mapped to zero. 
Each perturbed parameter \(\overline{\bm{\theta}}_i \in \mathbb{R}^{2}\), satisfying \(\|\overline{\bm{\theta}}_i\| \leq l\), serves as a data point for least squares estimation. Define $\phi$ as the vector of second-order monomials:
\begin{align}
\phi(\bm{\theta})=[\tfrac{1}{2}\theta^1\theta^1,\theta^1\theta^2,\tfrac{1}{2}\theta^2\theta^2]^\top.
\end{align}
Then, the design matrix $\bm{\Phi}\in\mathbb{R}^{s\times 3}$, where $s$ is the number of such data points, is constructed as 
\begin{align}
\bm{\Phi} = \left[\phi(\overline{\bm{\theta}}_1)^\top, \ldots, \phi(\overline{\bm{\theta}}_{s})^\top\right]^\top.
\end{align}
Now, representing the symmetric Hessian matrix estimate $\widehat{\mathbf{H}}$ as a vector $\widehat{\mathbf{h}}=[\widehat{H}_{11},\widehat{H}_{12},\widehat{H}_{22}]^{\top}$, where $\widehat{H}_{ml}$ is the $(m,l)$-th element of $\widehat{\mathbf{H}}$, we can express the least square estimate $\widehat{\mathbf{h}}$ as:
\begin{align}
\label{e:lsq_supp}
\widehat{\mathbf{h}}&=\mathop{\rm arg\,min}\limits_{\mathbf{h}\in \mathbb{R}^3} \sum_{i=1}^{s}\left(q(\overline{\bm{\theta}}_i)-\phi(\overline{\bm{\theta}}_i)^\top\mathbf{h}\right)^2,\\
q(\overline{\bm{\theta}}_i)&=f(\overline{\bm{\theta}}_i)-\widehat{\mathbf{g}}^\top \overline{\bm{\theta}}_i-f(\bm{\theta}).
\end{align}
Here, we use $f(\bm{\theta})$ to denote $f(\mathbf{x}^k)$. The solution $\widehat{\mathbf{h}}$ has a closed-form expression:
\begin{align}
\widehat{\mathbf{h}}=(\bm{\Phi}^\top \bm{\Phi})^{-1}\bm{\Phi}^\top \mathbf{q}, 
\end{align}
where $\mathbf{q} =\left[q(\overline{\bm{\theta}}_1),\ldots,q(\overline{\bm{\theta}}_{s})\right]^\top$. Similarly to the estimated Hessian $\widehat{\mathbf{h}}$, let $\mathbf{h}$ represent the entries of $\mathbf{H}$. Then, their difference $\Delta \mathbf{h}$ can be expressed as
\begin{align}
\Delta\mathbf{h}
\;=\;
\widehat{\mathbf{h}} -\mathbf{h}
&
=\bigl(\bm{\Phi}^\top \bm{\Phi})^{-1}\bm{\Phi}^\top\Delta \mathbf{q},
\label{e:deltah}
\end{align}
where $\Delta\mathbf{q} = \mathbf{q} - \bm{\Phi}\mathbf{h}$.

The vector $\Delta \mathbf{q}$ captures the discrepancy between the true
function value $f(\overline{\boldsymbol{\theta}})$ and its second-order expansion, where the linear term is replaced by an estimated subspace gradient $\widehat{\mathbf{g}}$. Specifically, the $i$-th element $\Delta q_i$ of $\Delta\mathbf{q} \in \mathbb{R}^{s}$ is given by  
\begin{align}
\Delta q_i
&= f(\overline{\bm{\theta}}_{i})-f(\bm{\theta})-\widehat{\mathbf{g}}^{\top}\overline{\bm{\theta}}-\tfrac{1}{2}\,\overline{\bm{\theta}}_i^\top\,\mathbf{H}\,\overline{\bm{\theta}}_i\\
&= \left( f(\bm{\theta}) 
          + \nabla f(\mathbf{x}^k)^\top \mathbf{P}^{\top} \overline{\bm{\theta}}_{i}
          + \frac{1}{2}\,\overline{\bm{\theta}}_{i}^\top \mathbf{P} 
            \,H f(\bm{\theta}) \,\mathbf{g}^\top\,\overline{\bm{\theta}}_{i}
          + \mathcal{R}_i \right)
   \;-\;
   \left( f(\bm{\theta}) 
          + \widehat{\mathbf{g}}^{\top}\,\overline{\bm{\theta}}_i
          + \tfrac{1}{2}\,\overline{\bm{\theta}}_i^\top\,\mathbf{H}\,\overline{\bm{\theta}}_i \right)\\
&= \left(\mathbf{g}\,\nabla f(\mathbf{x}^k) - \widehat{\mathbf{g}}\right)^{\top}\,\overline{\bm{\theta}}_i + \mathcal{R}_i,
\label{e:deltaqbound}
\end{align}
where $\mathcal{R}_i$ denotes the Taylor series remainder term, accounting for the mismatch between the exact function value and its second-order approximation. By the Lipschitz continuity of the Hessian (Assumption~\ref{a:assumptions1_supp}), $\mathcal{R}_i$ is bounded 
as~\cite{NW06}:
\begin{align}
\label{e:rbound}
\mathcal{R}_i \leq \frac{C_{2}}{6}\|\mathbf{P}^{\top}\overline{\bm{\theta}}^{i}\|^{3} =  \frac{C_2}{6}\|\overline{\bm{\theta}}^{i}\|^{3} = \frac{C_2}{6}l^{3}.
\end{align}

To bound $\Delta \mathbf{h}$, we decompose it into a product of two norms:
\begin{align}
\|\Delta\mathbf{h}\| = \|(\bm{\Phi}^{\top}\bm{\Phi})^{-1}\bm{\Phi}^{\top}\Delta\mathbf{q}\| \leq \|(\bm{\Phi}^{\top}\bm{\Phi})^{-1}\|_{2}\|\bm{\Phi}^{\top}\Delta\mathbf{q}\|.
\end{align}
The square of the norm $\|\bm{\Phi}^{\top}\Delta\mathbf{q}\|$ can be explicitly written as:
\begin{align}
\bm{\Phi}^{\top} \Delta\mathbf{q} &=
\left[\sum_{i=1}^{s}\frac{1}{2} \theta_{i}^{1}\theta_{i}^{1} \Delta q_i, 
\sum_{i=1}^{s} \theta_{i}^{1}\theta_{i}^{2} \Delta q_i, 
\sum_{i=1}^{s} \frac{1}{2}\theta_{i}^{2}\theta_{i}^{2} \Delta q_i
\right]^\top,\\
\left\|\bm{\Phi}^{\top} \Delta\mathbf{q}\right\|^2 &=
\left(\sum_{i=1}^{s}\frac{1}{2} \theta_{i}^{1}\theta_{i}^{1} \Delta q_i\right)^2+\left(\sum_{i=1}^{s} \theta_{i}^{1}\theta_{i}^{2} \Delta q_i\right)^2 
+\left(\sum_{i=1}^{s} \frac{1}{2}\theta_{i}^{2}\theta_{i}^{2} \Delta q_i\right)^2.
\end{align}

To bound this term, we first bound \( \Delta q_i \) and the monomial coefficients. Using \cref{e:gradientbound}, the first summand in $\Delta q_i$ (\cref{e:deltaqbound}) is bounded as 
\begin{align}
\label{e:deltaqfirst}
 \left({\mathbf{P}}\nabla f(\mathbf{x}^{k}) -\widehat{\mathbf{g}}\right)^{\top}\overline{\boldsymbol{\theta}}_{i} &\leq \|{\mathbf{P}}\nabla f(\mathbf{x}^{k}) -\widehat{\mathbf{g}}\|\|\overline{\boldsymbol{\theta}}_{i}\| \leq \frac{C_1}{\sqrt{2}}\epsilon^{k} l.
\end{align}
Combining \cref{e:deltaqfirst} and \cref{e:rbound}, we obtain a bound on $\Delta q_{i}$:
\begin{align}
\|\Delta q_{i}\| 
&\leq \left\| \mathbf{P}^{\top} \nabla f(\mathbf{x}^{k}) - \widehat{\mathbf{g}} \right\|\left\| \overline{\boldsymbol{\theta}}_{i} \right\| + \left\| \mathcal{R}_{i} \right\| \\
&\leq \frac{C_1}{\sqrt{2}} \epsilon^{k} l + \frac{C_2}{6} l^3.
\label{e:boundqi}
\end{align}
Since the magnitude of perturbation $\overline{\bm{\theta}}_i$ is at most $l$, it follows that
\begin{align}
(\theta_{i}^{1})^2,\; (\theta_{i}^{2})^2,\; |\theta_{i}^{1}\theta_{i}^{2}|\leq l^2.
\end{align}
Using this and the bound on $\Delta q_i$ from \cref{e:boundqi}, we obtain:
\begin{align}
\left\|\bm{\Phi}^{\top} \Delta\mathbf{q}\right\|^2 
&= \left(\left(\sum_{i=1}^{s}\frac{1}{2} \theta_{i}^{1}\theta_{i}^{1}\right)^2 +\left(\sum_{i=1}^{s}\theta_{i}^{1}\theta_{i}^{2}\right)^2 +\left(\sum_{i=1}^{s}\frac{1}{2} \theta_{i}^{2}\theta_{i}^{2}\right)^2\right) \left(\sum_{i=1}^{s}\Delta q_i\right)^2\\
&\leq 3 s^2l^4 \left(\frac{C_1}{\sqrt{2}} \epsilon^{k} l + \frac{C_2}{6} l^3\right)^2\\
\Leftrightarrow \left\|\bm{\Phi}^{\top} \Delta\mathbf{q}\right\|&\leq \sqrt{3} sl^2 \left(\frac{C_1}{\sqrt{2}} \epsilon^{k} l + \frac{C_2}{6} l^3\right).
\label{e:phideltaqbound}
\end{align}

Using Assumption~\ref{a:assumptions2_supp}, $\|(\bm{\Phi}^{\top}\bm{\Phi})^{-1}\|_{2}=\frac{1}{\gamma}$. Combining this with \cref{e:phideltaqbound}, we obtain:
\begin{align}
\|\Delta\mathbf{h}\|
&\leq \|(\bm{\Phi}^{\top}\bm{\Phi})^{-1}\|_{2}\|\bm{\Phi}^{\top}\Delta\mathbf{q}\| \\
&\leq \sqrt{3}\left( \frac{s}{\gamma}\, \left( \frac{C_1}{\sqrt{2}}\epsilon^{k}\,l^3 
+ \frac{C_2}{6}\,l^5 \right) \right).
\end{align}
Since the spectral norm of a matrix is always smaller than its Frobenius norm, we can now finally bound the error between the estimated subspace Hessian and its analytical counterpart as:
\begin{align}
\|\widehat{\mathbf{H}}-\mathbf{H}\|_{2} &\leq \sqrt{2}\|\Delta \mathbf{h}\| \\
&\leq \sqrt{6} \frac{s}{\gamma}\, \left( \frac{C_1}{\sqrt{2}}\epsilon^{k}\,l^3 
+ \frac{C_2}{6}\,l^5 \right).
\label{e:hbound}
\end{align}
\end{proof}

We assumed a lower bound $\gamma$ on the smallest eigenvalue of $\bm{\Phi}^\top \bm{\Phi}$. This assumption is not restrictive in our setting, as the radius $l$ of the sampled points $\{\overline{\bm{\theta}}_i\}_{i=1}^s$ does not need to decrease as $k$ grows. Our algorithm and the convergence proof do not require $\|\widehat{\mathbf{H}}-\mathbf{H}\|_{2}\to 0$ as $k$, which would otherwise necessitate $l\to 0$ for more precise estimation. In applications where such precision is needed, the number of sampled points $s$ should increase toward infinity to ensure a lower-bounded eigenvalue~\cite{AT07}. While analyzing this setting would be theoretically appealing, it is not necessary for our application.

\subsection{Proof of Proposition~\ref{p:mainresult_supp}}
Here, we prove the convergence of our algorithm, which uses the \emph{Armijo backtracking line search}~\cite{NW06} with a minimum step size. We begin by explicitly representing the update vector $\mathbf{v}^k$:
\begin{align}
\mathbf{v}^k=\sum_{j=1}^{\frac{m}{2}}P^{-1}\left((\widehat{\mathbf{H}}_j^k)^{-1}\widehat{\mathbf{g}}_j^k\right),
\end{align} where $m$ is the dimension of the intermediate subspace $\mathcal{V}$.

To ensure that $\widehat{\mathbf{H}}$ remains positive definite, eigenvalue clipping is applied, replacing each eigenvalue $\lambda_i$ with $\min(\lambda_i, \kappa)$, where the positive parameter $\kappa$. As a preliminary step, we derive an upper bound for the subspace Hessian. 
\begin{align}
\label{e:hessian_bound}
\|\widehat{\mathbf{H}}\|_{2}=\|\widehat{\mathbf{H}}-\mathbf{P} H f(\mathbf{x}^k) \mathbf{P}^{\top}+\mathbf{P} H f(\mathbf{x}^k)\mathbf{P}^{\top}\|_{2} &\leq \|\widehat{\mathbf{H}}-\mathbf{P} H f(\mathbf{x}^k)\mathbf{P}^{\top}\|_{2}+\|\mathbf{P} H f(\mathbf{x}^k) \mathbf{P}^{\top}\|_{2} \\\
&\leq
\|\widehat{\mathbf{H}}-\mathbf{H}\|_{2} +\|\mathbf{P}\|_{2} \|H f(\mathbf{x}^k)\|_{2} \|\mathbf{P}\|_{2}.
\label{e:hessian_bound2}
\end{align}
By the Lipschitz continuity of the gradient $\nabla f$ (Assumption~\ref{a:assumptions1_supp}), the Hessian $H f(\mathbf{x}^k)$ is bounded: $\|H f(\mathbf{x}^k)\|_2 \leq C_1$~\cite{Fe23}.

Using this and noting that $\|\widehat{\mathbf{H}}\widehat{\mathbf{H}}^{-1}\|_{2} \leq \|\widehat{\mathbf{H}}\|_{2}\|\widehat{\mathbf{H}}^{-1}\|_{2}$, we derive a lower bound of $\|\widehat{\mathbf{H}}^{-1}\|_{2}$ from \cref{e:hessian_bound2}:
\begin{align}
  \|\widehat{\mathbf{H}}^{-1}\|_{2} \geq \frac{1}{\|\widehat{\mathbf{H}}\|_{2}} \geq \frac{1}{\|\widehat{\mathbf{H}}-\mathbf{H}\|_{2} + C_1+\kappa}.
\label{e:lowerbound}
\end{align}

Next, we establish a bound on the estimated subspace gradients in scenarios where $\mathcal{W}$ determined through random coordinate selection. Following the approach in~\cite{RPA25}, we take the expectation with respect to the projection matrices $\mathbf{P}$, which map onto the respective subspaces. This yields
\begin{align}
    \mathbb{E}_{\mathbf{P} \sim \mathcal{D}} [\|\mathbf{P}^{\top}\widehat{\mathbf{g}} -\nabla f(\mathbf{x}^{k})\|]  
    &= \mathbb{E}_{\mathbf{P} \sim \mathcal{D}}[\|\mathbf{P}^{\top}\widehat{\mathbf{g}}-\mathbf{P}^{\top}\mathbf{P}\nabla f(\mathbf{x}^{k})  
    + \mathbf{P}^{\top}\mathbf{P}\nabla f(\mathbf{x}^{k})-\nabla f(\mathbf{x}^{k})\|] \\
    &\leq \mathbb{E}_{\mathbf{P} \sim \mathcal{D}}[\|\mathbf{P}^{\top}\|_{2}\|\widehat{\mathbf{g}}-\mathbf{P}\nabla f(\mathbf{x}^{k})\|
    + \|\mathbf{P}^{\top}\mathbf{P}\nabla f(\mathbf{x}^{k})-\nabla f(\mathbf{x}^{k})\|] \\
    &\leq \frac{C_1\epsilon^{k}}{\sqrt{2}} + \left(1-\frac{2}{d}\right)\|\nabla
    f(\mathbf{x}^{k})\|.
\label{e:grad}
\end{align}
Given that $\epsilon^k < \frac{\sqrt{2}}{dC_1}\|\nabla f(\mathbf{x}^{k})\|$ (Assumption~\ref{a:assumptions2_supp}), there exists a constant $\tau^k$ satisfying $0< \tau^k <1$, such that~\cite{WQY24}:
\begin{align}
\label{e:tau}
    \frac{C_1\epsilon^{k}}{\sqrt{2}} + \left(1-\frac{2}{d}\right)\|\nabla f(\mathbf{x}^{k})\| &= \tau^{k} \|\nabla f(\mathbf{x}^{k})\| < \left(1-\frac{1}{d}\right)\|\nabla f(\mathbf{x}^{k})\| .
\end{align}
Combining~\cref{e:grad} and~\cref{e:tau}, we obtain
\begin{align}
\label{e:grad_inequal}
 \mathbb{E}_{\mathbf{P} \sim \mathcal{D}} [\|\mathbf{P}^{\top}\widehat{\mathbf{g}} -\nabla f(\mathbf{x}^{k})\|]  \leq \tau^{k} \|\nabla f(\mathbf{x}^{k})\|.
\end{align}
Applying the triangular inequality,
\begin{align}
\label{e:gradient_bound}
        \mathbb{E}_{\mathbf{p} \sim \mathcal{D}}[(1-\tau^k)\|\nabla f(\mathbf{x}^{k})\|] \leq \mathbb{E}_{\mathbf{g} \sim \mathcal{D}}[\|\mathbf{g}^{\top}\widehat{\mathbf{g}}\|] \leq \mathbb{E}_{\mathbf{p} \sim \mathcal{D}}[(1+\tau^k)\|\nabla f(\mathbf{x}^{k})\|].
\end{align}

Using Assumption~\ref{a:assumptions2_supp} and combining~\cref{e:hessian_bound} with~\cref{e:gradient_bound}, we obtain the iterative bound:
\begin{equation}
\label{e:iterative_gradient_inequal}
 \mathbb{E}_{\mathbf{g} \sim \mathcal{D}}[f(\mathbf{x}^{k+1})] \leq \mathbb{E}_{\mathbf{g} \sim \mathcal{D}}[f(\mathbf{x}^{k})]-c_1\rho^{\min}\frac{m(1-\tau^k)^{2}}{2(C_{1}+\|\widehat{\mathbf{H}}-\mathbf{H}\|_{2}+\kappa)^{2}}\|\nabla f(\mathbf{x}^{k})\|^{2}.
\end{equation}
Further, the Lipschitz continuity of the gradient provides an upper bound on $\epsilon^k$:
\begin{align}
 \epsilon^{k} < \frac{\sqrt{2}}{dC_1}\|\nabla f(\mathbf{x}^{k})\| < \frac{\sqrt{2}}{d}\Delta_{0},
\end{align}
where $\Delta_0 = \|\mathbf{x}^{0}-\mathbf{x}^{*}\|$. Substituting this bound into the upper bound of $\|\widehat{\mathbf{H}}-\mathbf{H}\|_{2}$ in \cref{e:hbound}, we obtain

\begin{align}
\|\widehat{\mathbf{H}}-\mathbf{H}\|_{2}\leq E:=
      \sqrt{6}\frac{s}{\gamma}\,
      \left(
         \frac{C_{1}}{d}\Delta_{0}l^{3} 
         \;+\;
         \frac{C_{2}}{6}l^{5}
      \right).
\end{align}
From~\cref{e:tau}, we establish that $\tau^k<1-\frac{1}{d}$, which further implies 
\begin{align}
\frac{1}{d^2} <(1-\tau^k)^{2}.
\end{align}

By substituting the results from~\cref{e:tau} and~\cref{e:grad_inequal} into~\cref{e:iterative_gradient_inequal}, we obtain the following bound:
\begin{equation}
\label{e:iterative_expect_bound}
 \mathbb{E}_{\mathbf{g} \sim \mathcal{D}}[f(\mathbf{x}^{k+1})] \leq \mathbb{E}_{\mathbf{g} \sim \mathcal{D}}[f(\mathbf{x}^{k})]-c_1\rho^{\min}\frac{m}{2d^{2}(C_{1}+E+\kappa)^{2}}\mathbb{E}_{\mathbf{g} \sim \mathcal{D}}[\|\nabla f(\mathbf{x}^{k})\|^{2}].
\end{equation}

\noindent Using the $\mu$-strong convexity of $f$ (Assumption~\ref{a:assumptions1_supp}) and the Polyak--{\L}ojasiewicz inequality~\cite{KNS16}, we obtain:
\begin{align}
\|\nabla f(\mathbf{x}^{k})\|^{2} \geq 2\mu(f(\mathbf{x}^{k})-f(\mathbf{x}^{*})).
\label{e:polyak}
\end{align}
Substituting~\cref{e:polyak} into~\cref{e:iterative_expect_bound}, we get
\begin{align}
\label{e:max_bound_of_expect}
 \mathbb{E}_{\mathbf{g} \sim \mathcal{D}}[f(\mathbf{x}^{k+1})] \leq \mathbb{E}_{\mathbf{g} \sim \mathcal{D}}[f(\mathbf{x}^{k})]-c_1\rho^{\min}\frac{m\mu}{d^{2}(C_{1}+E+\kappa)^{2}}\mathbb{E}_{\mathbf{g} \sim \mathcal{D}}[f(\mathbf{x}^{k})-f(\mathbf{x}^{*})].
\end{align}
Subtracting $f(\mathbf{x}^{*})$ from both sides of~\cref{e:max_bound_of_expect}, we obtain
\begin{align}
\label{e:subtract_expect_bount}
 \mathbb{E}_{\mathbf{g} \sim \mathcal{D}}[f(\mathbf{x}^{k+1})] - \mathbb{E}_{\mathbf{g} \sim \mathcal{D}}[f(\mathbf{x}^{*})] &\leq \mathbb{E}_{\mathbf{g} \sim \mathcal{D}}[f(\mathbf{x}^{k})]- \mathbb{E}_{\mathbf{g} \sim \mathcal{D}}[f(\mathbf{x}^{*})]-c_1\rho^{\min}\frac{m\mu}{d^{2}(C_{1}+E+\kappa)^{2}}\mathbb{E}_{\mathbf{g} \sim \mathcal{D}}[f(\mathbf{x}^{k})-f(\mathbf{x}^{*})]\\
 &=(1-\nu)\{\mathbb{E}_{\mathbf{g} \sim \mathcal{D}}[f(\mathbf{x}^{k})]-\mathbb{E}_{\mathbf{g} \sim \mathcal{D}}[f(\mathbf{x}^{*})]\},
\end{align}
where
\begin{align}
 \nu &= c_1\rho^{\min}\cfrac{m\mu}{d^{2}(C_1+E+\kappa)^{2}}.
\end{align}

Performing the telescoping multiplication from $0$ to $k-1$, as in~\cite{GKL19}, we obtain
\begin{align}
\mathbb{E}_{\mathbf{g} \sim \mathcal{D}}[f(\mathbf{x}^{k+1})  - f(\mathbf{x}^{*})]  \leq (1-\nu)^{k}(f(\mathbf{x}^{0})-f(\mathbf{x}^{*})).
\label{e:telemul}
\end{align}
The Lipschitz continuity of the gradient further provides the bound:
\begin{align}
f(\mathbf{x}^{0}) \leq f(\mathbf{x}^{*}) + \nabla f(\mathbf{x}^{*})^{\top}(\mathbf{x}-\mathbf{x}^{*})+\frac{C_1}{2}\|\mathbf{x}-\mathbf{x}^{*}\|^{2}.
\end{align}
\noindent Since $\nabla f(\mathbf{x}^{*}) = 0$, we have
\begin{align}
 f(\mathbf{x}^{0}) -f(\mathbf{x}^{*}) \leq \frac{C_1}{2}{\Delta_{0}}^{2}.
 \label{e:zerosum}
\end{align}

\noindent Finally, substituting~\cref{e:zerosum} to~\cref{e:telemul}, we obtain
\begin{align}
\mathbb{E}_{\mathbf{g} \sim \mathcal{D}}[f(\mathbf{x}^{k})-f(\mathbf{x}^{*})] \leq (1-\nu)^{k}\frac{C_1}{2}{\Delta_{0}}^{2}.
\end{align}

\section{Additional information}
\label{s:expdetails}

\subsection{Preliminaries for zeroth-order optimization} 
\label{s:preliminaries}
\paragraph{First- and second-order optimization.} Consider the problem of minimizing a twice-continuously differentiable function $f:\mathbb{R}^d\mapsto\mathbb{R}$. Two well-established strategies for this task are first-order and second-order iterative descent methods. At each iteration $k$, these methods update the current \emph{search point} $\mathbf{x}^k$ using the following update rules: 
\begin{align}
\label{e:firstorderopt_supp}
\mathbf{x}^{k+1}&=\mathbf{x}^k-\eta\nabla f(\mathbf{x}^k),&& \text{\small{(first-order method)}}\\
\mathbf{x}^{k+1}&=\mathbf{x}^k-\eta(\mathbf{b}^k)^{-1}\nabla f(\mathbf{x}^k),&&\text{\small{(second-order method)}}
\label{e:secondorderopt_supp}
\end{align}
where $\eta$ is the step size (or learning rate), and $\mathbf{B}^k$ is a positive definite (PD) matrix that influences the search direction and scaling. A representative example of a second-order method is Newton's method, where $\mathbf{B}^k$ corresponds to the Hessian matrix $\mathbf{H}^k:=Hf(\mathbf{x}^k)$, with $H$ denoting the Hessian operator. 

\paragraph{Zeroth-order optimization.} 
A common approach in ZO optimization is to approximate the gradient $\nabla f(\mathbf{x})$ using randomized finite difference methods while adhering to the first-order update rule (\cref{e:firstorderopt})~\cite{NS15,DJW15}:
\begin{equation}
\label{e:rge_supp}
\widehat{\nabla}f(\mathbf{x}) = \frac{1}{q}\sum_{i=1}^{q} \frac{f(\mathbf{x}+\epsilon \mathbf{u}_{i})-f(\mathbf{x})}{\epsilon}\mathbf{u}_{i},
\end{equation}
where $\epsilon$ is a small positive scalar, $q$ is the number of perturbations, and $\mathbf{u}_i\in \mathbb{R}^d$ is a random perturbation vector drawn from the normal distribution $\mathcal{N}(\mathbf{0},\mathbf{I})$.

\noindent For $q = 1$ and $\epsilon \rightarrow 0$, this approximation converges to the directional derivative of $f$ at $\mathbf{x}$ along $\mathbf{u}$~\cite{NS15}:
\begin{align}
D_\mathbf{u} f(\mathbf{x}) = \mathbf{u}^{\top}\nabla f(\mathbf{x}) = \lim_{\epsilon \rightarrow 0}\frac{f(\mathbf{x}+\epsilon\mathbf{u})-f(\mathbf{x})}{\epsilon}.
\end{align}
Multiplying by $\mathbf{u}$ and taking the expectation with respect to $\mathbf{u}$ yields:
\begin{align}
\mathbb{E}[D_\mathbf{u} (\mathbf{x})\mathbf{u}] = \mathbb{E}[(\mathbf{u}\mathbf{u}^{\top})\nabla f(\mathbf{x})] = \nabla f(\mathbf{x}).  
\end{align}
The Hessian matrix can also be approximated using finite differences~\cite{Ye23}:
\begin{align}
\label{e:hessiafd}
\widehat{\mathbf{H}} = \frac{1}{q}\sum_{i=1}^{q}\frac{f(\mathbf{x}+\epsilon\mathbf{u}_i)+f(\mathbf{x}-\epsilon\mathbf{u}_i)-2f(\mathbf{x})}{2\epsilon^2}\mathbf{u}_{i}\mathbf{u}_{i}^{\top} + \lambda \mathbf{I},
\end{align}
where $\lambda$ is a regularization parameter that ensures $\widehat{\mathbf{H}}$ remains PD. This estimated Hessian can be combined with the finite-difference gradient $\widehat{\nabla}f$ (\cref{e:rge_supp}) to enable second-order updates within the ZO setting. However, in high-dimensional settings, the query complexity $q$ required for accurate Hessian estimation scales quadratically with the dimensionality $d$, making direct second-order ZO optimization impractical.

\subsection{Baseline algorithms}
\label{sss:baseline}
To benchmark our approach, we evaluated several existing zeroth-order optimization methods. 
\begin{itemize}
\item \emph{Randomized stochastic projected gradient} (\textbf{\texttt{RSPG}}) applies standard randomized finite differences to approximate function gradients~\cite{GLZ16}. 
\item \emph{Zeroth-order sign-based stochastic gradient descent} (\textbf{\texttt{ZO-signSGD}}) improves robustness by relying only on the directional signs of the estimated partial derivatives~\cite{LCC19}. 
\item Extending the adaptive momentum method (AdaMM) to the zeroth-order domain, \emph{zeroth-order adaptive momentum method} (\textbf{\texttt{ZO-AdaMM}}) accelerates optimization by updating learning rates and descent directions based on past gradient evaluations~\cite{CLX19}. 
\item \emph{Sample reusable linear interpolation-based zeroth-order optimization} (\textbf{\texttt{ReLIZO}}) computes gradient approximations using quadratically constrained linear programming on sampled function evaluations~\cite{WQY24}. 
\item \emph{Gradient compressed sensing} (\textbf{\texttt{GraCe}}) reduces function evaluation costs by calculating sparse approximate gradients~\cite{QT24}. 
\item \emph{Hessian informed zeroth-order optimizer} (\textbf{\texttt{HiZOO}}) exploits second-order curvature information by estimating the diagonal elements of the Hessian matrix~\cite{ZDY25}. 
\item \emph{Zeroth-order Hessian-aware algorithm} (\textbf{\texttt{ZOHA}}) estimates the Hessian matrix using low-rank approximations via eigenvalue decomposition~\cite{Ye23}.  
\item \emph{Zeroth-order optimization for deep model training} (\textbf{\texttt{DeepZero}}) is designed for neural network training, selectively optimizing specific layers at each step~\cite{CZJ24}. When applied to logistic regression, interpreted as a neural network with a single layer, it simplifies to randomized block coordinate descent.
\end{itemize}

\subsection{Datasets} 
\label{sss:dataset}
Following the experimental setups of existing ZO optimization studies~\cite{LCC19,ctl22,LZG24}, we conducted experiments on standard logistic regression tasks using widely used benchmark datasets. The dataset statistics for our experiments are summarized in~\Cref{t:logregdata}.  

\begin{table}[hbt!]
  \centering
  \begin{tabular}{@{}ccc@{}}
    \toprule
    Dataset & Feature dimensions & Number of examples \\
    \midrule
    \emph{a3a}  & 123 & 3,186 \\ 
    \emph{a5a}   & 123 & 6,414 \\
    \emph{w1a}   & 300 & 2,477 \\
    \emph{GISETTE}   & 5,000 & 6,000 \\
    \emph{Olivetti} & 4,097  & 400 \\    
    \emph{phishing} & 68 & 11,055 \\
    \bottomrule
  \end{tabular}
  \caption{Summary of datasets used for logistic regression experiments.}
  \label{t:logregdata}
\end{table}

\subsection{Experimental details}
\label{sss:implinfo}
\paragraph{Logistic regression.}
For \textbf{\texttt{ZO-AdaMM}}, we set the hyperparameters $\beta_1 = 0.9$ and $\beta_2 = 0.5$ as specified in the original paper~\cite{CLX19}. For \textbf{\texttt{ReLIZO}}, we configured the reuse bound $b$ to $0.05 \eta$, where $\eta$ is the step size, and applied a maximum reuse rate of 50\%, following the recommendations in~\cite{WQY24}. Each optimization step used a function evaluation budget of 10, consistent with experimental settings in~\cite{WQY24}. For \textbf{\texttt{GraCe}}, we set the sparsity level $s$ to 5, used a single repeat ($m = 1$), and applied a scaling factor $\gamma = 0.7$, based on the best-performing configuration reported in~\cite{QT24}. For \textbf{\texttt{HiZOO}}, we used a smoothing factor $\alpha = 10^{-6}$, which showed the best performance in~\cite{ZDY25}. 
Finally, we applied backtracking line search to all algorithms, setting the initial step size to 1.0, the Armijo constant to $c_1 = 10^{-4}$, and the shrinkage factor to $0.5$.

\paragraph{Neural networks.}
For all experiments, we initialized the initial learning rate at $0.1$ and gradually decreased it using a cosine decay scheduler. A weight decay of $5 \times 10^{-4}$ was consistently applied throughout the 50 training epochs. In our algorithm, \textbf{\texttt{ZO-SAH}}, the subspaces $\mathcal{V}$, $\{\mathcal{W}\}$ were updated every 20 steps ($T = 20$). Following~\cite{CZJ24}, $\mathcal{V}$ was determined using \textbf{\texttt{ZO-GraSP}}, a ZO adaptation of the GraSP pruning method~\cite{CGR20}, which randomly samples projection dimensions with probabilities inversely proportional to the absolute value of the corresponding partial derivatives of $f$.

\section{Additional experiments}
\label{s:extendedresults}

\subsection{Extended results}
\label{sss:extended}
\begin{figure*}[h!]
    \centering
    \begin{minipage}{\linewidth}
        \begin{subfigure}{0.495\linewidth}
            \centering
            \includegraphics[width=\linewidth]{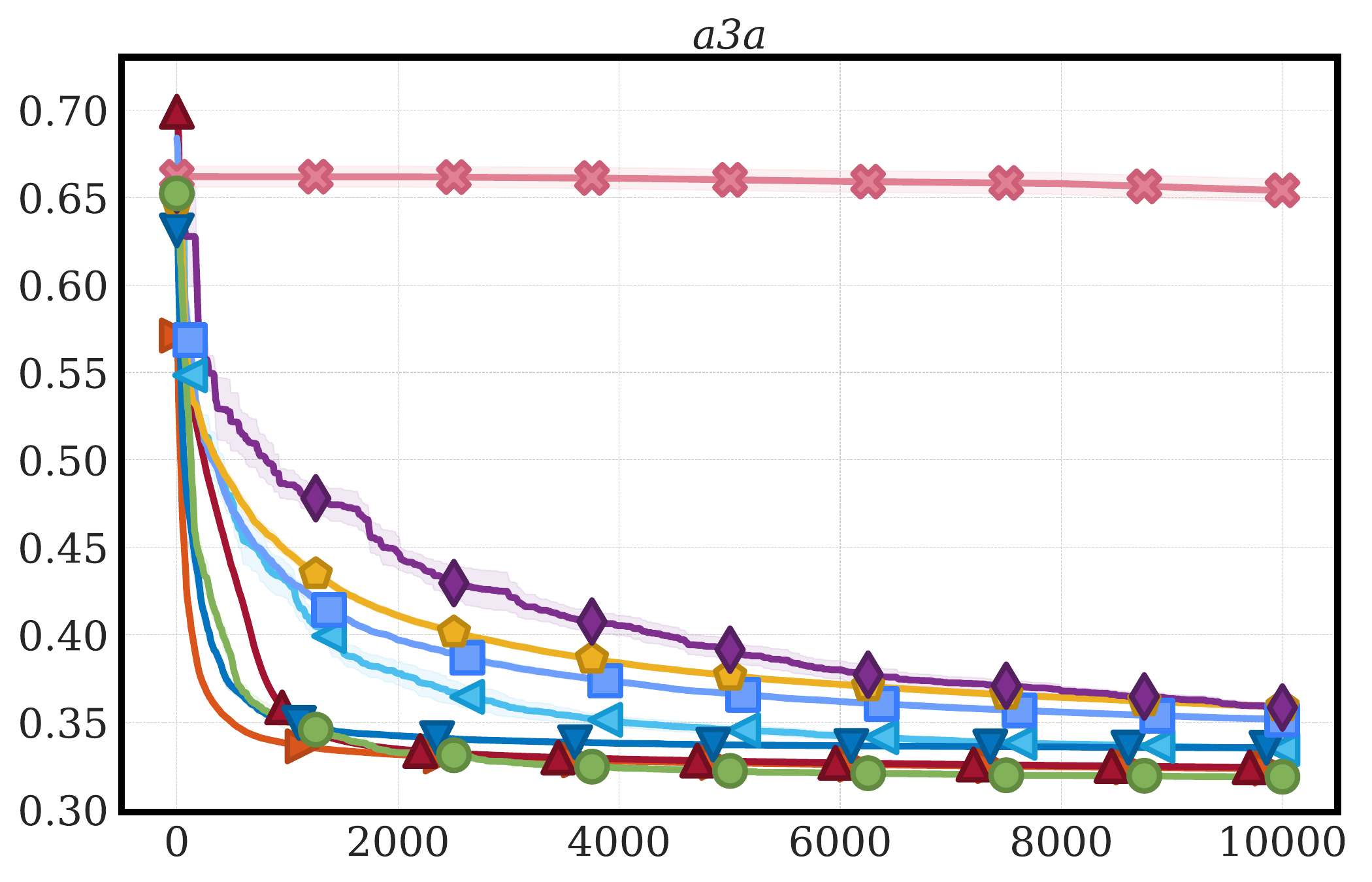} 
            \label{sf:a3a_10000}
        \end{subfigure}
        \begin{subfigure}{0.495\linewidth}
            \centering
            \includegraphics[width=\linewidth]{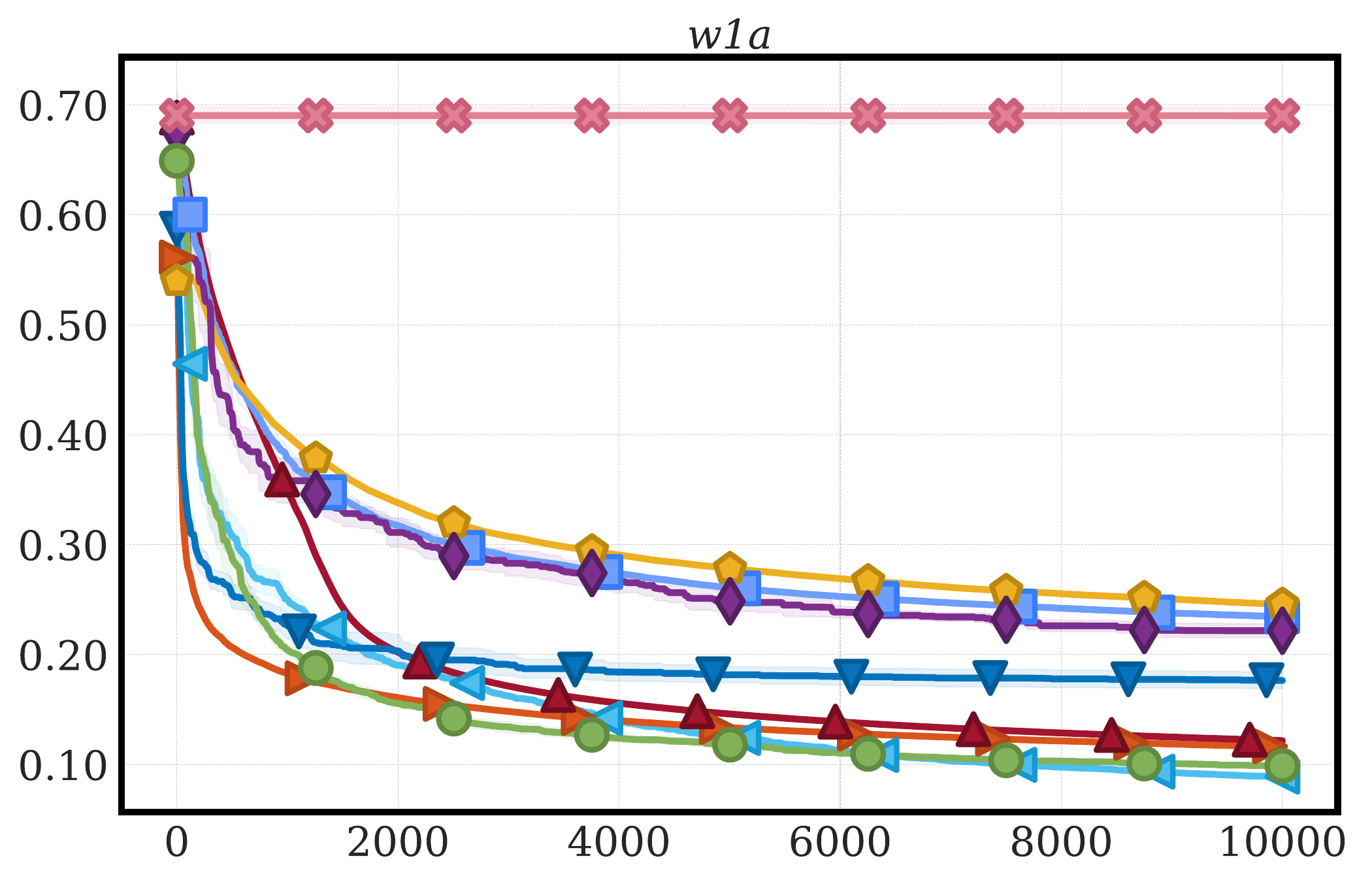} 
            \label{sf:w1a_10000}
        \end{subfigure}
    \end{minipage}
    \vspace{-10pt}    

    \begin{minipage}{\linewidth}
    \begin{subfigure}{0.495\linewidth}
            \centering
            \includegraphics[width=\linewidth]{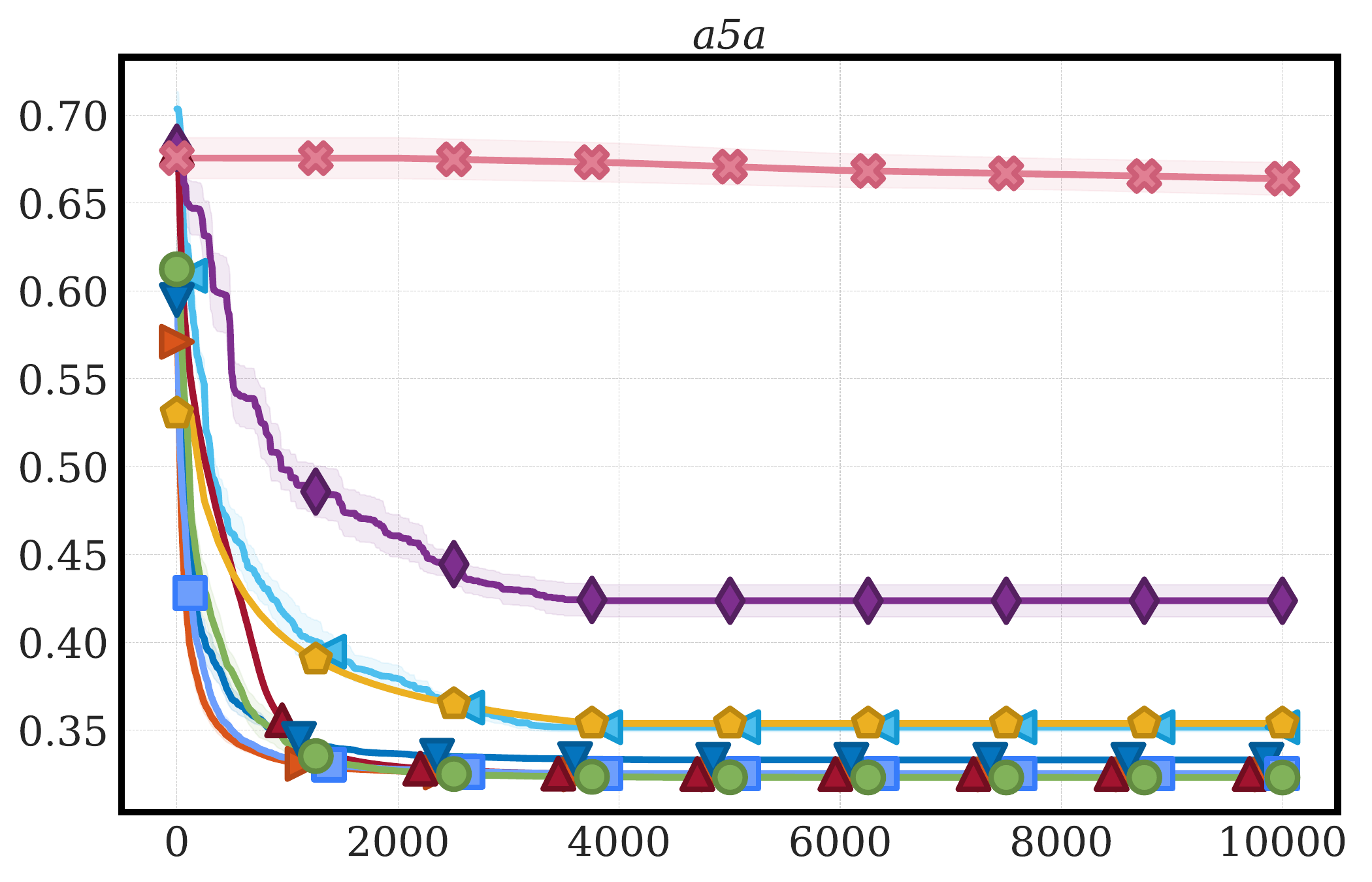} 
            \label{sf:a5a_10000}
        \end{subfigure}  
    \begin{subfigure}{0.495\linewidth}
            \centering
            \includegraphics[width=\linewidth]{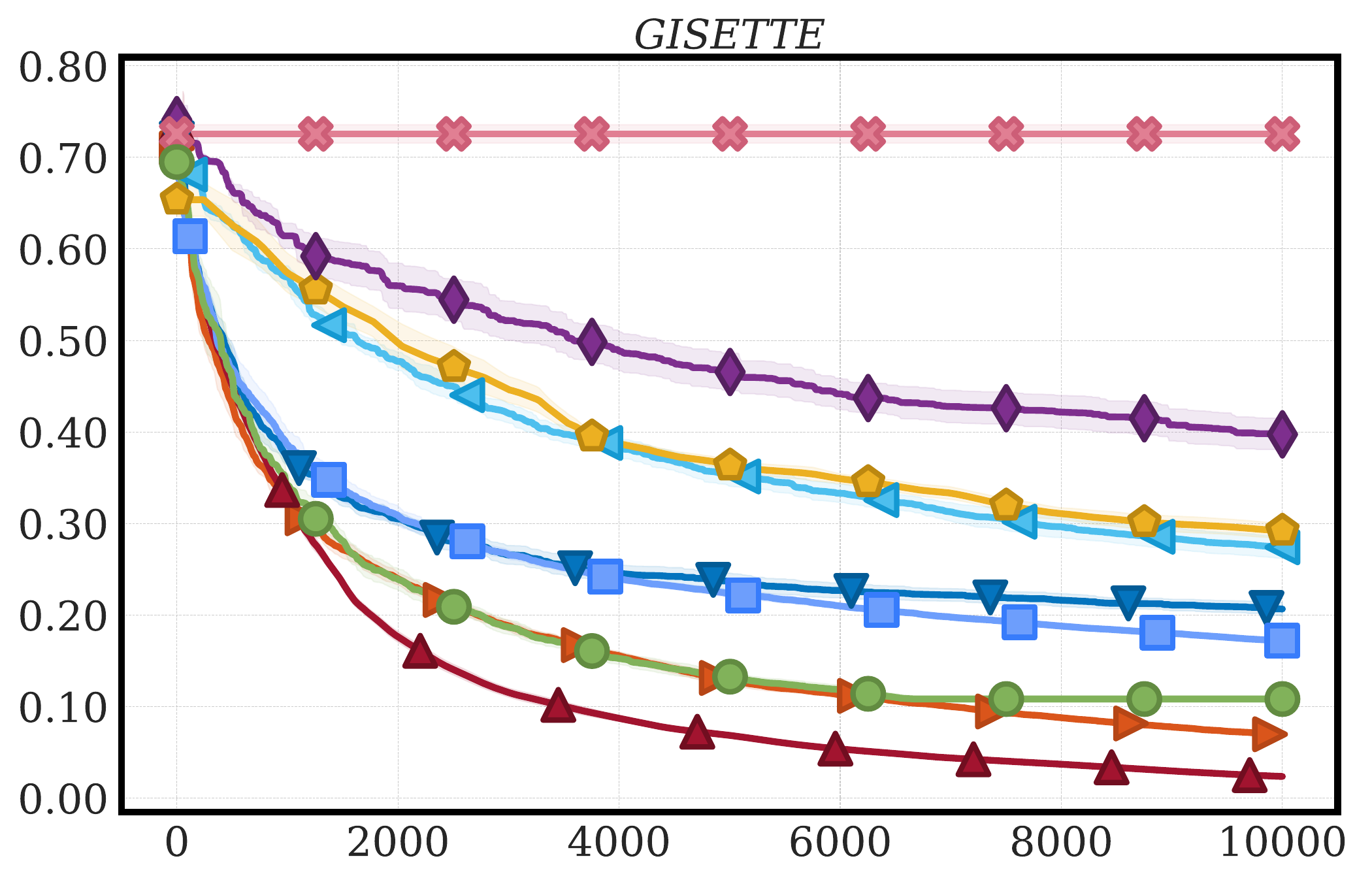} 
            \label{sf:gis_10000}
        \end{subfigure}        
    \end{minipage}
    \vspace{-10pt}

    \begin{minipage}{\linewidth}
        \begin{subfigure}{0.495\linewidth}
            \centering
            \includegraphics[width=\linewidth]{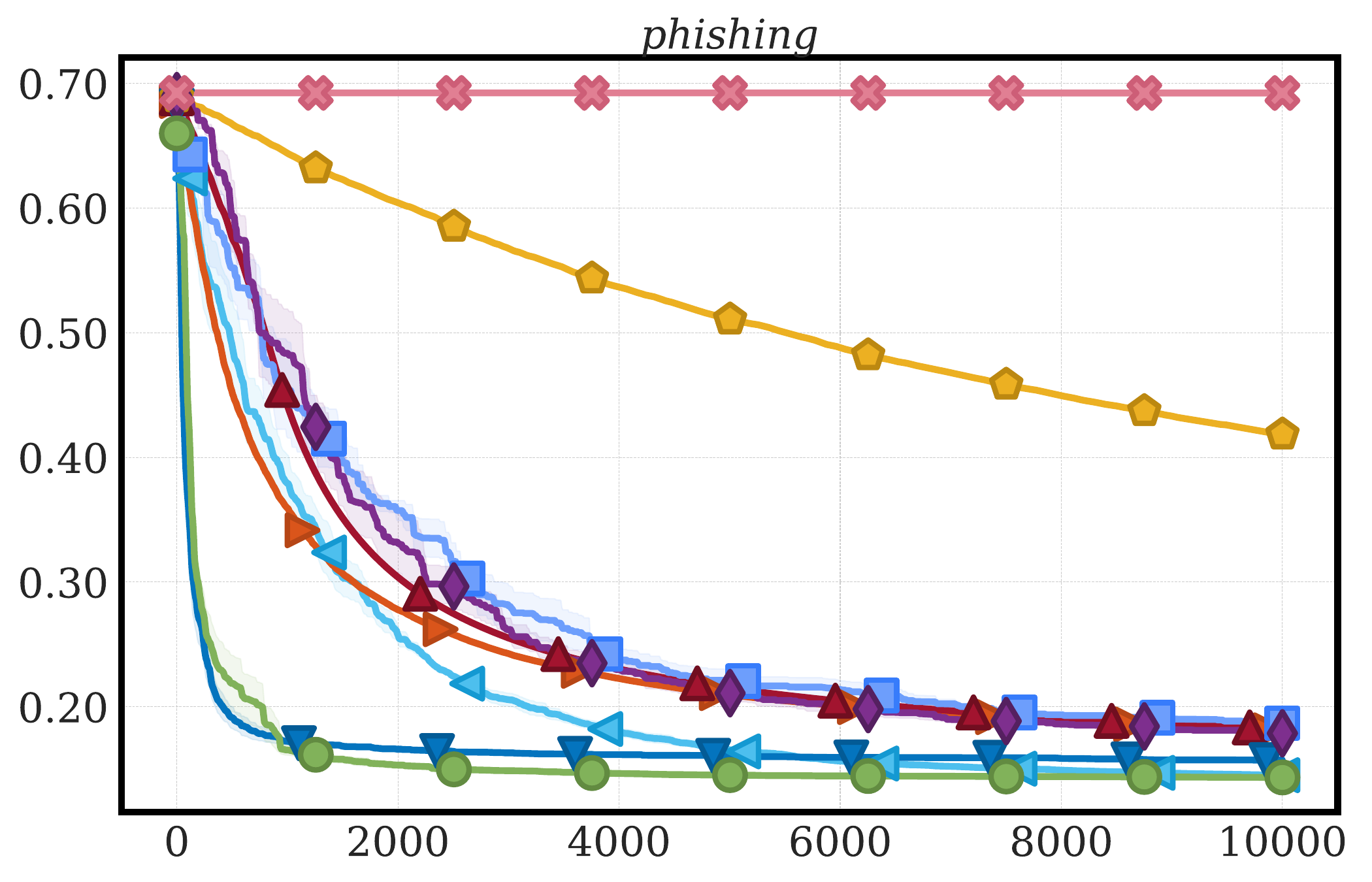} 
            \label{sf:phi_10000}
        \end{subfigure}
        \begin{subfigure}{0.495\linewidth}
            \centering
            \includegraphics[width=\linewidth]{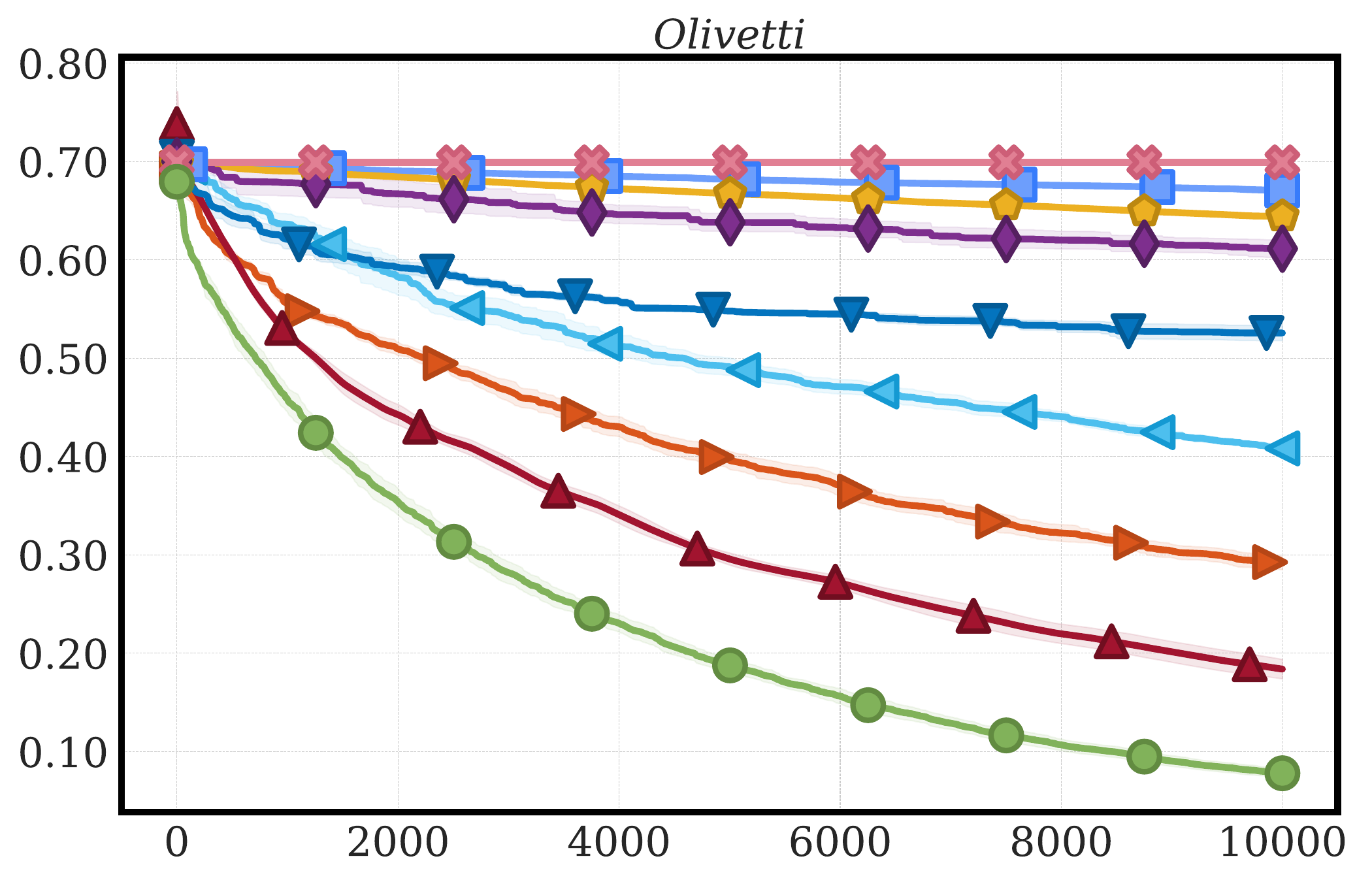} 
            \label{sf:oli_10000}
        \end{subfigure}  
    \end{minipage}
    \vspace{-10pt}

    \begin{subfigure}{\linewidth}
        \raggedleft
        \includegraphics[width=0.96\linewidth]{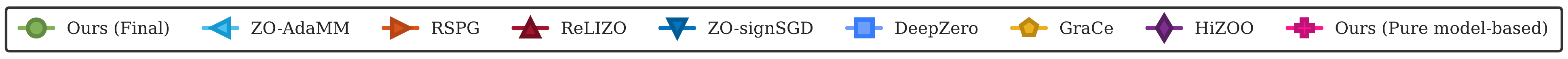}
    \end{subfigure}
    \caption{Progression of average objective function values (y-axis) across six logistic regression benchmark datasets. Since each algorithm step requires a different number of function evaluations, the x-axis is aligned to show the cumulative number of function evaluations. The starting x-coordinates of individual plots vary slightly, as they correspond to a single complete update step of each algorithm. Shaded regions represent twice the standard deviation.}
    \label{f:main_10000}
\end{figure*}

In \cref{s:logistic_result}, we presented logistic regression results for up to 5,000 function evaluations. \Cref{f:main_10000} extends this analysis, showing the convergence behavior over 10,000 function evaluations. We also include result for \textbf{\texttt{ZOHA}}. \textbf{\texttt{ZOHA}} is not specifically designed for function evaluation-efficient optimization scenarios: Each step of \textbf{\texttt{ZOHA}} requires 4$d$ function evaluations, where $d$ is the dimensionality of the optimization function $f$. For the \emph{Olivetti} and \emph{GISETTE} datasets, a single step of \textbf{\texttt{ZOHA}} exceeds 10,000 function evaluations, causing its convergence curve to remain flat at its initial value. Overall, the extended function evaluation results are consistent with the findings presented in \cref{s:logistic_result}.

\subsection{Additional ablation study}
\label{sss:ablation}
\begin{figure}[hbt!]
    \centering
    \begin{minipage}{0.495\columnwidth}
        \centering
        \includegraphics[width=\columnwidth]{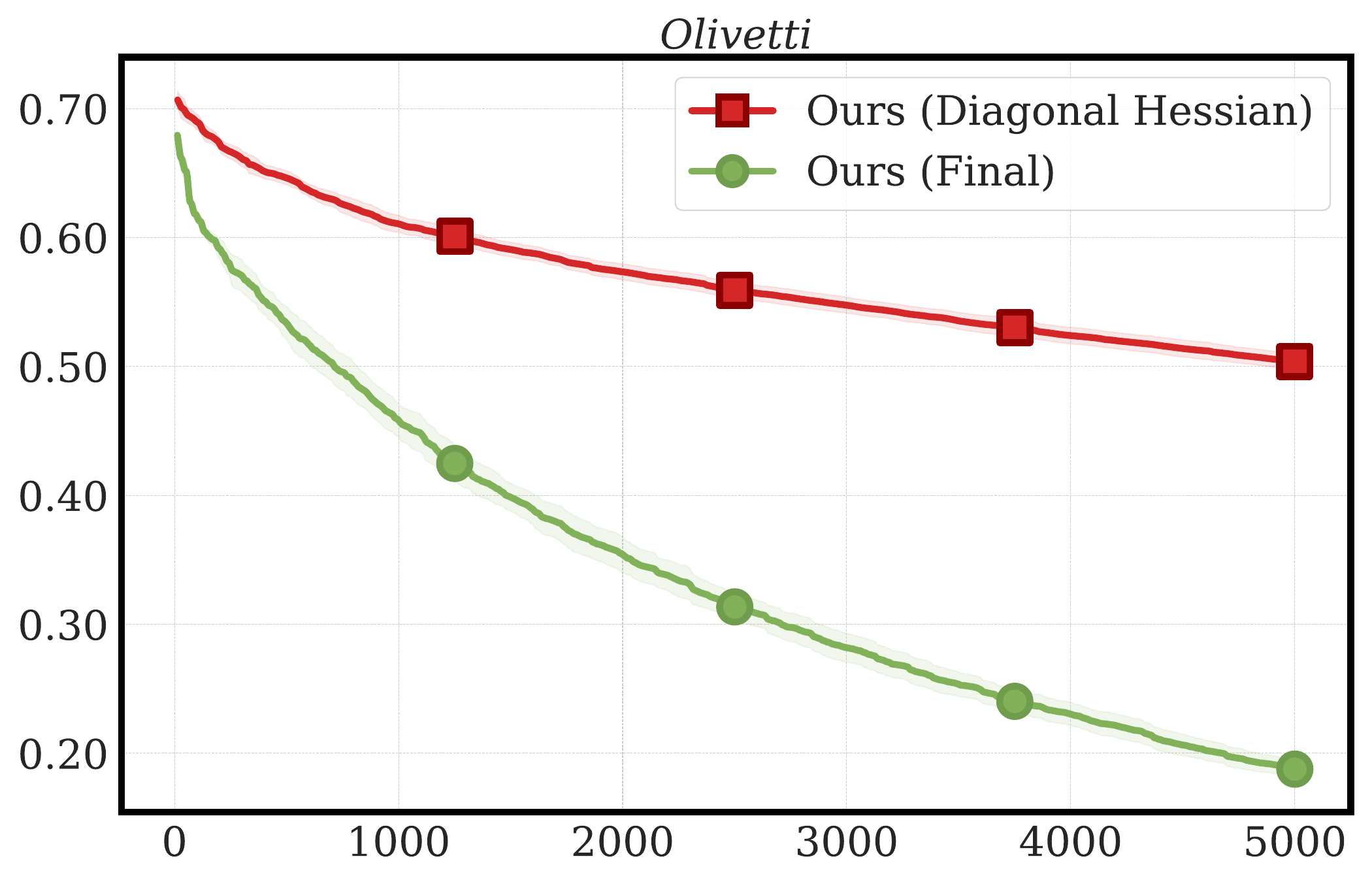}
        \label{sf:lr_diag}
    \end{minipage}
    \begin{minipage}{0.495\columnwidth}
        \centering
        \includegraphics[width=\columnwidth]{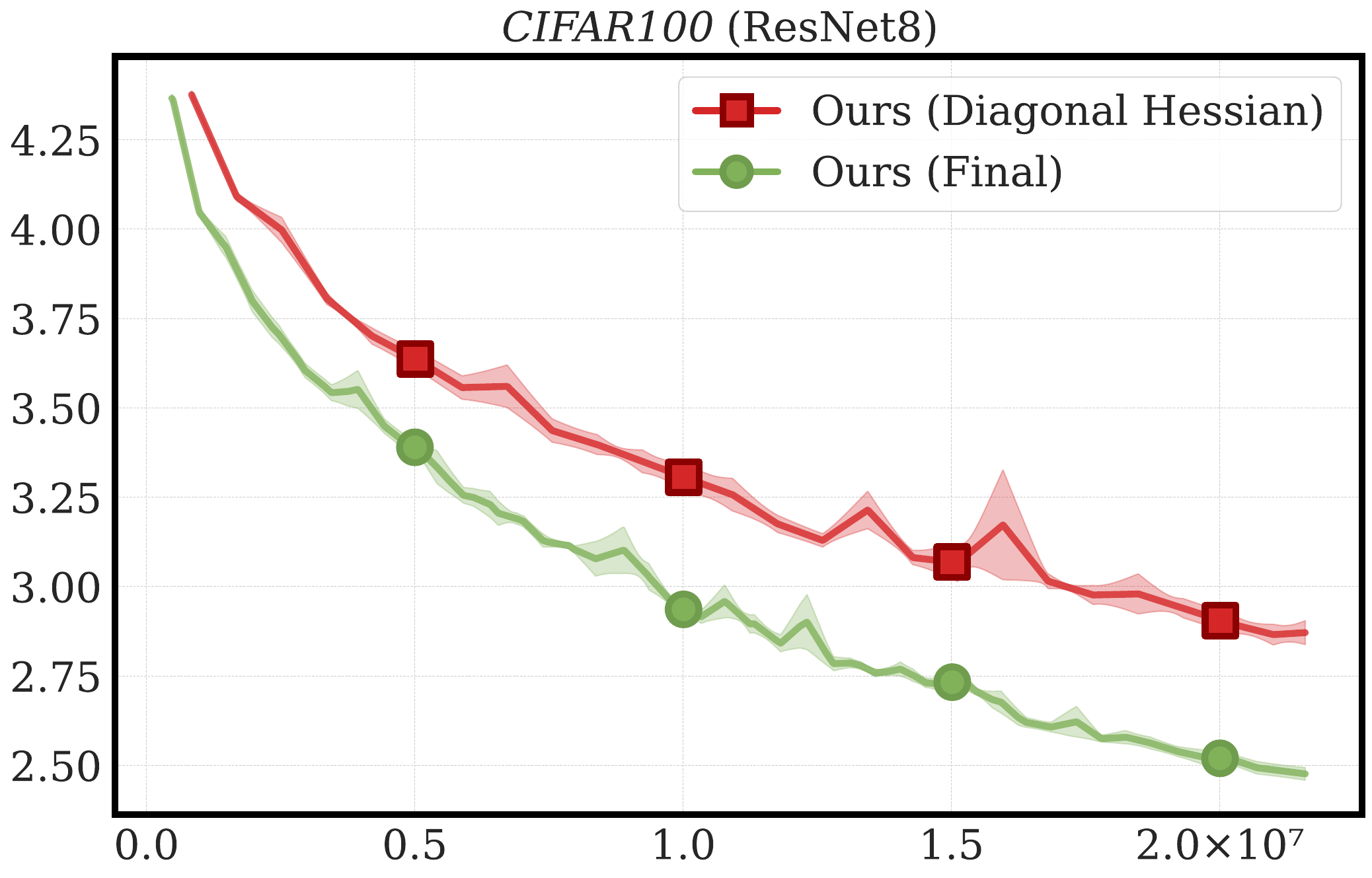}
        \label{sf:nn_diag}
    \end{minipage}
    \vspace{-6mm}
    \caption{Impact of incorporating off-diagonal Hessian elements: (left) logistic regression and (right) neural network training.}
    \label{f:ablationdiagonal}
\end{figure}

\paragraph{Comparison with diagonal only Hessian estimates.}
Our algorithm exploits the full Hessian in the subspace $\mathcal{W}$. \Cref{f:ablationdiagonal} shows the performance of a variation of our algorithm that estimates and uses only the diagonal elements of the Hessian matrix (calculated using finite differences). The results emphasize the significance of incorporating off-diagonal Hessian elements, as they capture interactions between input variables in the function. This observation aligns with the findings in the \cref{s:logistic_result}, where our algorithm outperformed \textbf{\texttt{HiZOO}}, which also relies solely on diagonal Hessian elements.

\begin{figure}[htbp]
    \centering
    \begin{minipage}{0.495\linewidth}
        \centering
        \includegraphics[width=\linewidth]{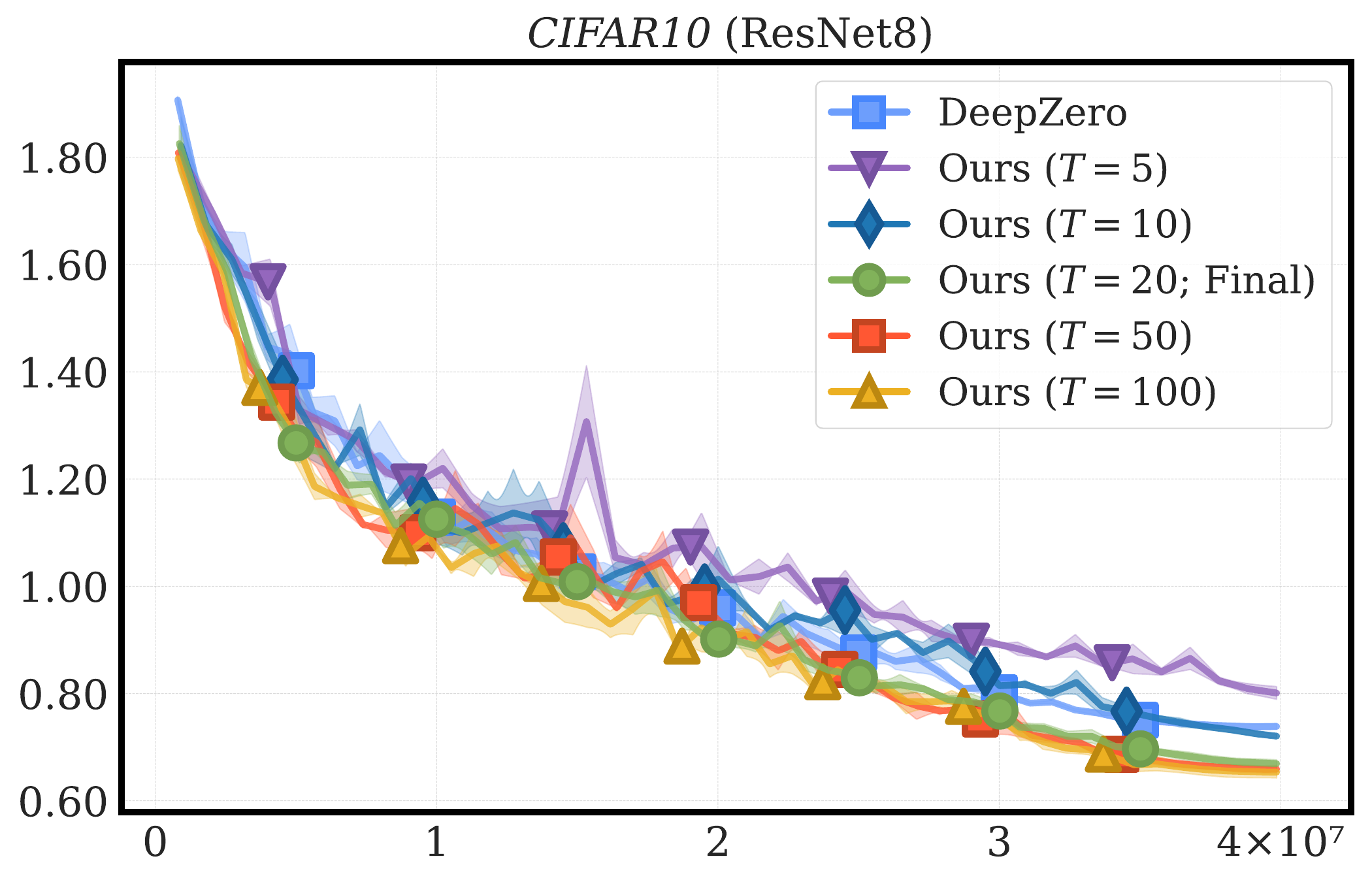}
        \label{sf:res8_10_T}
    \end{minipage}
    \begin{minipage}{0.495\linewidth}
        \centering
        \includegraphics[width=\linewidth]{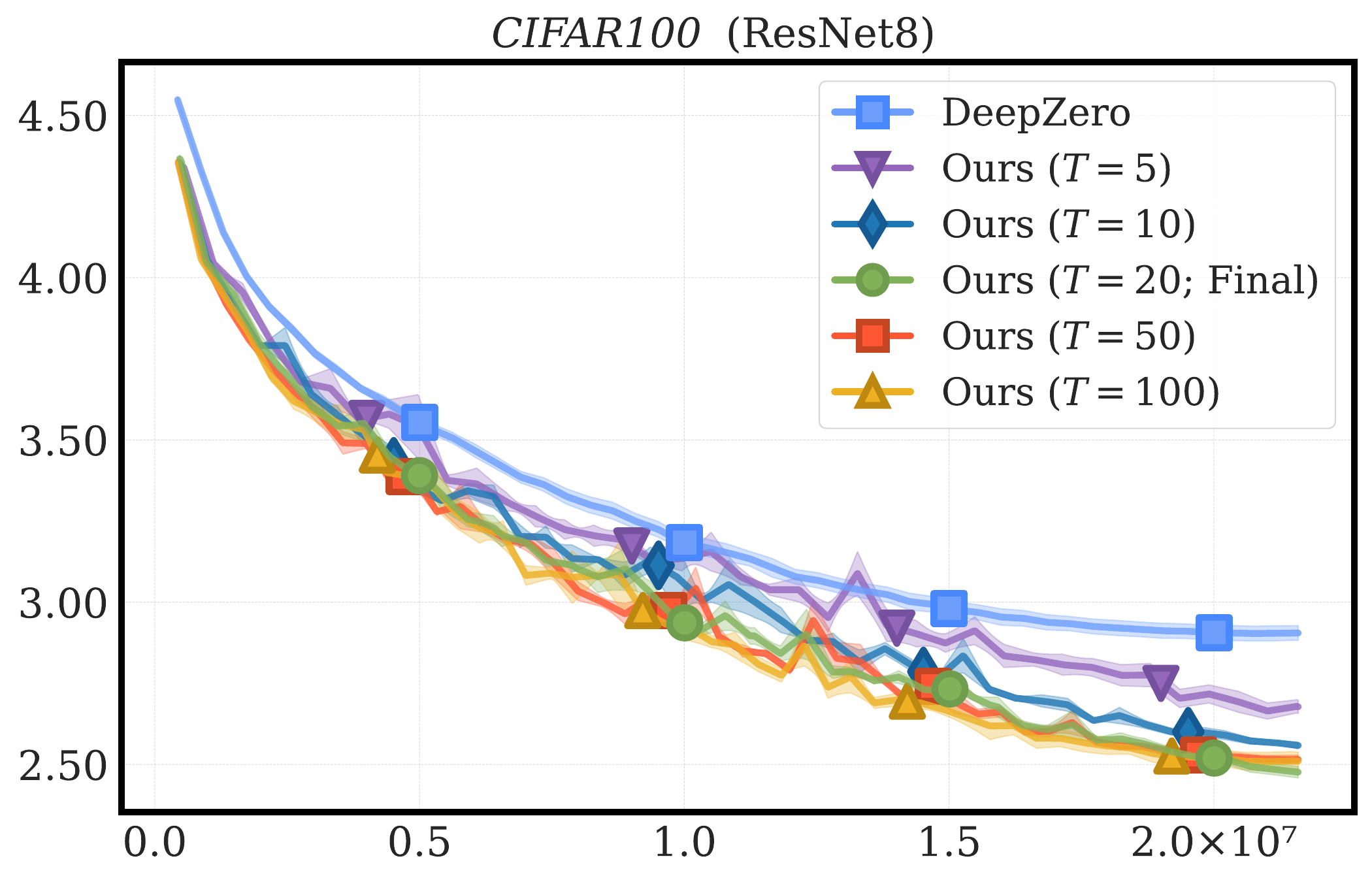}
        \label{sf:res8_100_T}
    \end{minipage}
    \vspace{-6mm}
    \caption{Performance of our algorithm with varying subspace switching period $T$ values and \textbf{\texttt{DeepZero}}.}
    \label{f:tvariation}
\end{figure}

\paragraph{Impact of subspace switching period $T$.} The switching period $T$ controls subspace updates and function reuse. A small $T$ leads to frequent subspace changes, increasing function evaluations and negating caching benefits. Conversely, a larger $T$ promotes function reuse but may slow adaptation to the optimization landscape. We set $T=20$ to balance frequent subspace updates with efficient function reuse (See~\cref{f:tvariation,t:iteration}).

\begin{table}[hbt!]
    \centering
    \begin{tabular}{ccccccc}
    \hline
    \multirow{2}{*}{Model} & \multicolumn{5}{c}{$T$} \\ 
    \cline{2-6}
    & 5 & 10 & 20 & 50 & 100 \\
    \hline
    \emph{CIFAR10} (ResNet8) & 0.80$\pm$0.01 & 0.72$\pm$0.00 & 0.67$\pm$0.01 & 0.66$\pm$0.01 & 0.66$\pm$0.01 \\
    \emph{CIFAR100} (ResNet8) & 2.68$\pm$0.02 & 2.56$\pm$0.01 & 2.47$\pm$0.01 & 2.51$\pm$0.01 & 2.50$\pm$0.03 \\
    \hline
    \end{tabular}
    \caption{Final $f$ values of our algorithm for different $T$ values.}
    \label{t:iteration}
\end{table}

\begin{figure}[hbt!]
    \centering
    \begin{subfigure}{0.5\linewidth}
       \centering
       \includegraphics[width=\linewidth]{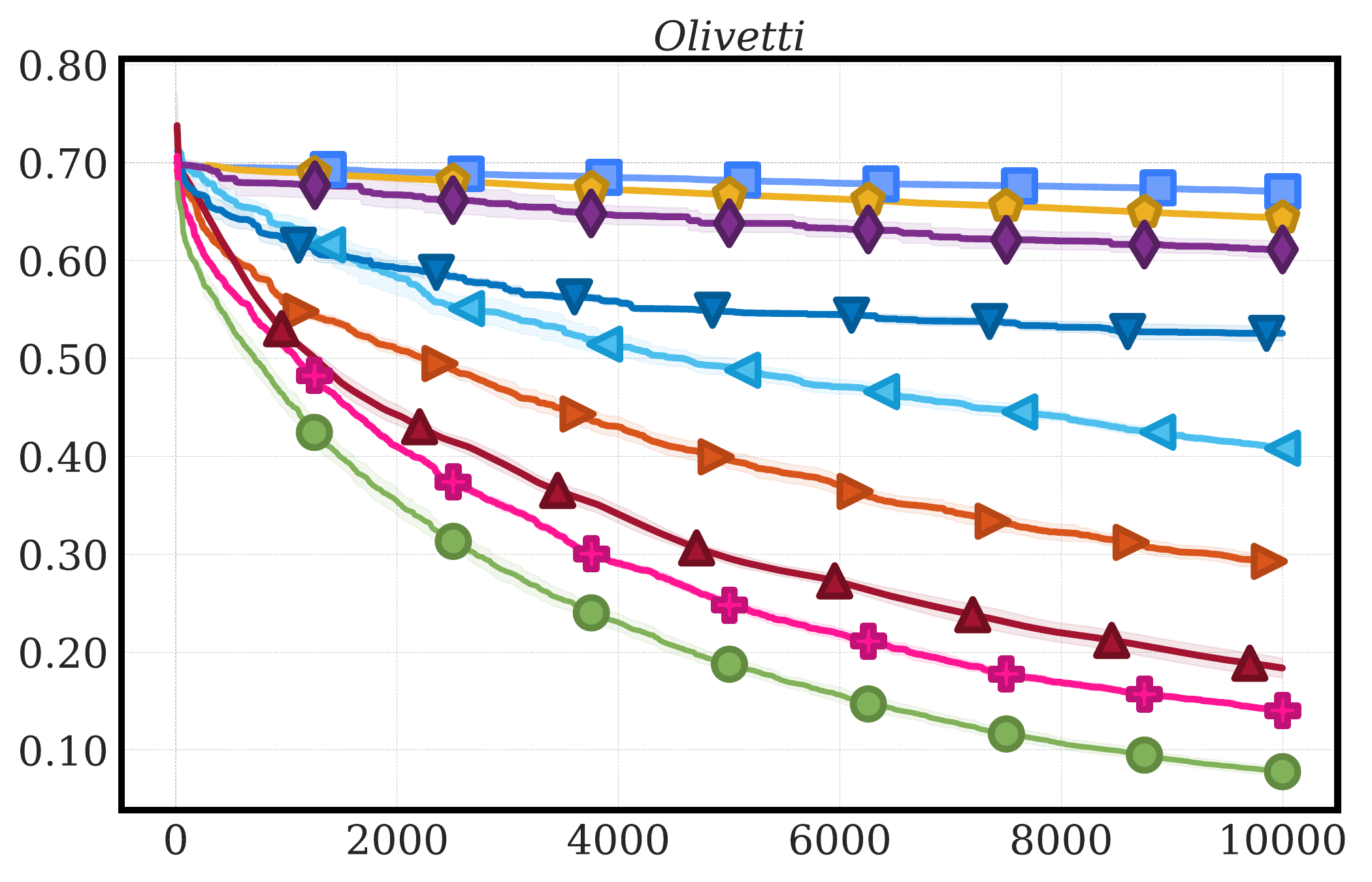} 
       \label{sf:interpo}
   \end{subfigure}
   \begin{subfigure}{\linewidth}
       \centering
       \includegraphics[width=\linewidth]{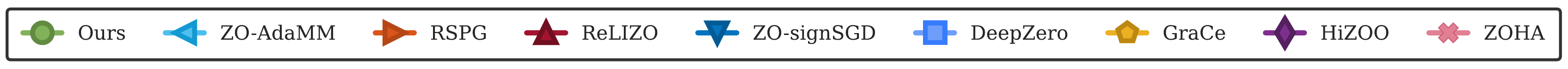} 
       \vspace{-10pt}
   \end{subfigure}
    \caption{Performance of the proposed \textbf{\texttt{ZO-SAH}} algorithm (Ours (Final)) and its variant using polynomial model-based estimation for both the gradient and Hessian (Ours (Pure model-based)). For reference, the convergence curves of other algorithms are also visualized.}
    \label{f:ablationmodelgradient}
\end{figure}

\paragraph{Comparison with a variation that estimates the gradient using model fitting.} Our algorithm constructs the Hessian estimate $\widehat{\mathbf{H}}$ by fitting a quadratic polynomial model of $f$, while the gradient estimate $\widehat{\mathbf{g}}$ is obtained using coordinate-wise finite differences. Although it is theoretically possible to estimate the gradient through model fitting as well, \cref{f:ablationmodelgradient} demonstrates that our final design leads to improved convergence over the purely model-based estimation of both the gradient and Hessian.

\end{document}